\documentclass[twocolumn]{fairmeta}

\usepackage{microtype}
\usepackage{graphicx}
\usepackage{subcaption}
\usepackage{booktabs} 
\usepackage{csquotes}

\usepackage{amsmath}
\usepackage{amssymb}
\usepackage{mathtools}
\usepackage{amsthm}
\usepackage{xspace}
\usepackage{colortbl}
\usepackage{multirow}
\usepackage{enumitem}
\usepackage{wrapfig}

\newcommand{\putalg}{{\small V-JEPA}\xspace}
\definecolor{fbApp}{HTML}{c8e7fa}
\definecolor{fbPurple3}{HTML}{f0ebf5}
\newcommand{\cc}{\cellcolor{metabg}}
\newcommand{\ccg}{\cellcolor{metabg}}

\definecolor{citecolor}{HTML}{0071BC}
\definecolor{linkcolor}{HTML}{ED1C24}

\theoremstyle{plain}

\theoremstyle{definition}

\theoremstyle{remark}

\usepackage[textsize=tiny]{todonotes}

\title{Revisiting Feature Prediction for Learning Visual Representations from Video}

\author[1,2,3]{Adrien Bardes}
\author[1,4]{Quentin Garrido}
\author[3,5,6]{Jean Ponce}
\author[1]{Xinlei Chen}
\author[1]{Michael Rabbat}
\author[1,5,6]{Yann LeCun}
\author[1,\dagger]{Mahmoud Assran}
\author[1,\dagger]{Nicolas Ballas}

\affiliation[1]{FAIR at Meta}
\affiliation[2]{Inria}
\affiliation[3]{\'Ecole normale sup\'erieure, CNRS, PSL Research University}
\affiliation[4]{Univ.~Gustave Eiffel, CNRS, LIGM}
\affiliation[5]{Courant Institute, New York University}
\affiliation[6]{Center for Data Science, New York University}

\contribution[\dagger]{Joint last author}

\abstract{This paper explores feature prediction as a stand-alone objective for unsupervised learning from video and introduces \putalg, a collection of vision models trained solely using a feature prediction objective, without the use of pretrained image encoders, text, negative examples, reconstruction, or other sources of supervision.
The models are trained on 2 million videos collected from public datasets and are evaluated on downstream image and video tasks.
Our results show that learning by predicting video features leads to versatile visual representations that perform well on both motion and appearance-based tasks, without adaption of the model's parameters; e.g., using a frozen backbone. Our largest model, a {\small ViT-H/16} trained only on videos, obtains {\small $81.9\%$} on Kinetics-400, {\small $72.2\%$} on Something-Something-v2, and {\small $77.9\%$} on ImageNet1K.
}

\date{\today}
\correspondence{\email{\{abardes, massran, ballasn\}@meta.com}}

\metadata[Code]{\url{https://github.com/facebookresearch/jepa}}
\metadata[Blogpost]{\href{https://ai.meta.com/blog/v-jepa-yann-lecun-ai-model-video-joint-embedding-predictive-architecture/}{Click here}}

\begin{document}
\maketitle

\section{Introduction}
Humans possess the remarkable ability to map low-level signals originating from the retina into a semantic spatio-temporal understanding of the world; synthesizing notions such as objects and global motion~\citep{spelke1995object}.
A long-standing goal of the machine learning community is to identify the principles or objectives that may guide such unsupervised learning in humans~\citep{field1994goal, berkes2005slow, hinton1989connectionist}.
One related hypothesis is based on the \emph{predictive feature principle}~\citep{rao1999predictive}, which posits that representations of temporally adjacent sensory stimuli should be predictive of each other.
\begin{figure}[t!]
    \vspace{-4mm}
    \centering
    \includegraphics[width=\linewidth]{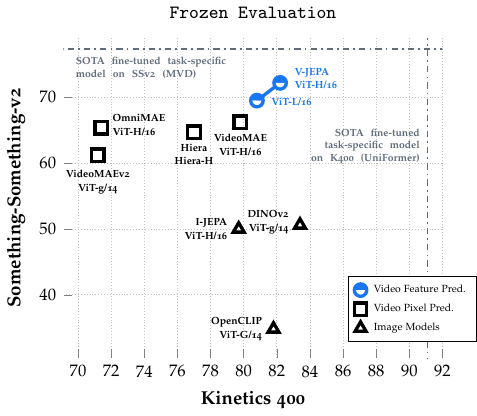}
    \caption{\putalg models pretrained on video learn versatile visual representations. It performs well on motion-based tasks (Something-Something-v2) and appearance-based tasks (Kinetics 400) without adaptation of the model's parameters, i.e., using the same frozen backbone for both tasks.}
    \label{fig:main_results}
\end{figure}

In this work, we revisit feature prediction as a stand-alone objective for unsupervised learning of visual representations from video.
Numerous advances in the field --- such as the standard use of transformer architectures in vision~\citep{dosovitskiy2020image}, the maturing of masked autoencoding frameworks~\citep{xie2021simmim, bao2021beit, he2021masked}, query-based feature pooling~\citep{chen2022context}, joint-embedding predictive architectures (JEPA)~\citep{lecun2022path, assran2023self, baevski2022data2vec}, and larger datasets --- form a unique arsenal of tools, which we integrate in a modern and conceptually simple method, the \emph{video joint-embedding predictive architecture} or \putalg, which is based solely on feature prediction, without using pretrained image encoders, text, negative examples, human annotations, or pixel-level reconstruction.

We seek to answer the simple question:
\begin{displayquote}\it
    How effective is feature prediction as a stand-alone objective for unsupervised learning from video with modern tools?
\end{displayquote}
To that end, we pretrain a family of \putalg models on a dataset of 2 million videos collected from publicly available datasets by combining a masked modeling prediction task with a joint-embedding predictive architecture (see Figure~\ref{fig:jepa}).
We measure performance on several downstream image and video tasks, using both frozen evaluation and end-to-end fine-tuning.
Our findings suggest that feature prediction can indeed serve as an effective stand-alone objective for unsupervised learning from video, while using significantly shorter training schedules than pixel prediction methods. Specifically:
\begin{itemize}
    \itemsep0em 
    \item Feature prediction leads to versatile visual representations that perform well across downstream image and video tasks without adaption of the model's weights; i.e., using a frozen backbone. \putalg achieves the best performance among methods we consider (+6\% accuracy) on the SomethingSomething-v2 task, which requires fine-grained temporal understanding. \putalg is also competitive on tasks like Kinetics400, where appearance-based features are sufficient and hence state-of-the-art image models such as DINOv2 excel (Figure~\ref{fig:main_results} and Table~\ref{tb:large_results}).
    \item Models trained with feature prediction are superior to pixel prediction approaches under a frozen evaluation protocol (attentive probing) and are competitive with pixel prediction under full fine-tuning, while using significantly shorter training schedules (Tables~\ref{tb:pixel_comparison} and \ref{tb:large_results}).
    \item Models trained with feature prediction are more label-efficient than pixel prediction approaches. Decreasing the available number of labeled examples results in an increase in the performance gap between V-JEPA and pixel-reconstruction models (Table~\ref{tb:lowshot}).
\end{itemize}

\section{Related Works}

\paragraph{\bf Slow Features.}
One way to encourage temporally adjacent representations to be predictive of each other is to ensure that they vary slowly over time.
Early works targeting predictive features encouraged representations of individual video frames to be locally temporally invariant, while preventing representation collapse by using spectral methods, as in SFA~\citep{wiskott2002slow}, SSA~\citep{kayser2001extracting}, and Simulated Fixations~\citep{zou2012deep}.
More recently,~\citet{goroshin2015unsupervised, wang2010learning} train a siamese convolutional network to map the representations of two subsequent frames to the same point, while encouraging distant frames to have diverse representations via a pair-wise margin loss and a triplet loss, respectively.
Other works~\citep{oord2018representation, suris2021learning, feichtenhofer2021large} implement temporal invariance using noise-contrastive estimation~\citep{gutmann2012noise}.
Our exploration in this paper goes beyond temporal invariance and explores feature prediction using masked modeling.
\vspace{-2mm}

\paragraph{\bf Predictive Features.}
Going beyond local invariance, a family of works trains a predictor network to map the representation of a frame or clip at one time-step to a distinct representation at another time-step.
\citet{srivastava2015unsupervised, vondrick2016anticipating, wang2023masked} train such a video feature predictor network on top of a frozen pretrained image or video encoder.
Unfreezing the target feature extractor, several methods train the video encoder and the predictor network simultaneously, while preventing collapse by using a supervised action forecasting loss~\citep{girdhar2021anticipative}, or by using the representations of distant clips as negative samples in a contrastive loss~\citep{han2019video, han2020memory, tan2023multiscale}, often focusing on small convolutional encoders~\citep{han2019video, han2020memory}.
The idea of learning a representation by predicting missing information in feature space is also core to the  joint-embedding predictive architecture (JEPA)~\citep{lecun2022path}, which combines a siamese encoder with a predictor network.
JEPAs have been successfully instantiated in several modalities, such as with audio data~\citep{ baevski2022data2vec} and image data~\citep{zhou2021ibotyes,oquab2023dinov2,assran2023self}.
In this work, we extend this paradigm to video data by leveraging recent advances in self-supervised learning.
\vspace{-1mm}
\paragraph{\bf Advances in Self-Supervised Learning.}
The use of vision transformers~\citep{dosovitskiy2020image, li2022uniformer} has become standard practice in self-supervised learning with joint-embedding architectures~\citep{chen2021empirical,caron2021emerging,oquab2023dinov2,zhou2021ibotyes,assran2022masked}, and unlocked masked image modeling in pixel space by parameterizing the pixel decoder as a transformer with learnable mask tokens~\citep{dosovitskiy2020image, xie2021simmim, he2021masked, bao2021beit}, demonstrating a step-change in the representation quality of autoencoding methods~\citep{vincent2010stacked}.
This line of generative methods was subsequently extended to video data using spatio-temporal masking~\citep{tong2022videomae, feichtenhofer2022masked, wang2023videomae, kalluri2023flavr, gupta2023siamese}.
It was also recently shown that the representations of masked image autoencoders could be significantly improved by using learnable pooling mechanisms based on cross-attention~\citep{chen2022context}.
Finally, through careful selection of design choices, the non-contrastive collapse prevention strategy in BYOL~\citep{grill2020bootstrap} was recently made to work with image feature prediction methods~\citep{baevski2022data2vec, assran2023self}, which demonstrated the ability to learn representations that can be leveraged for various downstream tasks without relying on invariance to hand-crafted image transformations.

\paragraph{\bf Feature Prediction versus Pixel Reconstruction.}
Approaches that predict in pixel space must dedicate significant model capacity and compute to capture all the low-level detail in the visual input.
By contrast, approaches that predict in latent space have the flexibility to eliminate irrelevant or unpredictable pixel-level details from the target representation~\citep{vondrick2016anticipating}.
Predicting in representation space has been shown to lead to versatile representations that perform well across many downstream tasks through linear probing or low-shot adaptation~\citep{assran2023self,oquab2023dinov2,assran2022masked}, while demonstrating an efficiency gain during pretraining compared to pixel level reconstruction~\citep{assran2023self, baevski2022data2vec, baevski2022efficient}.
The works of~\citet{baevski2022efficient, baevski2022data2vec} additionally show that predicting in representation space results in competitive end-to-end fine-tuning performance in the image, audio and text domains.
In this work, we extend these findings to the video modality.

\section{Methodology: Video-JEPA}
\label{sec:methodology}

\begin{figure}[h]
    \centering
    \includegraphics[width=0.7\linewidth]{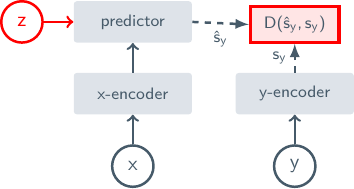}
    \caption{Joint-Embedding Predictive Architectures are trained to predict the representation of an input $y$ from the representation of another input $x$. The additional variable $z$ provides the predictor with information about the transformation that computes $y$ from $x$.}
    \label{fig:jepa}
\end{figure}
Our goal is to explore the effectiveness of feature prediction as a stand-alone objective for learning visual representations from video.
To that end, we use a joint-embedding predictive architecture (JEPA)~\citep{lecun2022path}; see Figure~\ref{fig:jepa}.
The main idea behind a JEPA is to learn by predicting the representation of an input $y$ from the representation of another input $x$.
The basic architecture is made up of an encoder, $E_\theta(\cdot)$, which computes the representation of the inputs, and a predictor, $P_\phi(\cdot)$, which predicts the representation of $y$ from the representation of $x$, conditioned on a variable $z$ indicating the transformation (or corruption) between $x$ and $y$.
Conditioning on $z$ enables the generation of distinct predictions for various transformations of $x$.

\subsection{Training Objective}
We train our visual encoder $E_\theta(\cdot)$ to satisfy the constraint that representations computed from one part of the video, $y$, should be predictable from representations computed from another part of the video, $x$.
The predictor network $P_\phi(\cdot)$, which maps the representation of $x$ to the representation of $y$, is trained simultaneously with the encoder, and is provided specification of the spatio-temporal positions of $y$ through the conditioning variable $z \gets \Delta_y$.

Naively implementing the objective using the regression
\[
    \text{minimize}_{\theta,\phi}\quad \lVert P_\phi(E_\theta(x), \Delta_y) - E_\theta(y) \rVert_1,
\]
would admit a trivial solution, where the encoder outputs a constant representation, regardless of its input.
In practice, we use the following modified objective to prevent representation collapse,
\begin{equation}
    \label{eq:loss}
    \text{minimize}_{\theta,\phi}\quad \rVert P_\phi(E_\theta(x), \Delta_y) - \text{sg}(\overline{E}_\theta(y)) \lVert_1,
\end{equation}
where $\text{sg}(\cdot)$ denotes a stop-gradient operation, which does not backpropagate through its argument, and $\overline{E}_\theta(\cdot)$ is an exponential moving average of the network $E_\theta(\cdot)$.
The use of an exponential-moving average feature extractor along with a stop-gradient and a predictor has been used as a collapse prevention strategy for image pretraining~\citep{grill2020bootstrap}, and studied empirically~\citep{xie2021simmim} and theoretically~\citep{tian2021understanding}.
In fact, the objective in equation~\eqref{eq:loss} is  similar to the loss of~\citet{assran2023self} used for image pretraining, but we modify it to use an $\ell_1$ regression, which we found to be more stable.

\paragraph{Theoretical motivation.}
A theoretical motivation for the effectiveness of this collapse prevention strategy was proposed in~\citet{grill2020bootstrap} for the BYOL method.
We provide a simple adaptation of their analysis for our $\ell_1$ loss.
For ease of exposition, we will disregard the effect of the conditioning variable $z$ and consider one dimensional representations.
Denote the representation $\overline{E}_\theta(y)$ by a random variable $Y$.
The optimal predictor under equation~\eqref{eq:loss} is thus given by the following functional expression,
\begin{align*}
    P^\star(E_\theta(x)) &= \text{argmin}_{P} \lVert P(E_\theta(x)) - Y \rVert_1\\
    &= \text{median}(Y|E_\theta(x)).
\end{align*}
Substituting this expression for the optimal predictor into the loss function and evaluating the expected gradient of the encoder gives
\[
     \nabla_\theta \mathbb{E} \lVert P^\star(E_\theta(x)) - Y \rVert_1 = \nabla_\theta  \text{MAD}(Y | E_\theta(x)),
\]
where $\text{MAD}(\cdot\ | E_\theta(x))$ is the median absolute deviation of a random variable conditioned on $E_\theta(x)$.
Thus, in the case where the predictor is optimal, the encoder must learn to capture as much information about the video as possible to minimize the deviation of the target.
The hypothesis is that incorporating an exponential moving average to compute the representation of $y$ ensures that the predictor evolves faster than the encoder and remains close to optimal, thereby preventing collapse.
\begin{figure*}[t]
    \centering
    \includegraphics[width=0.9\linewidth]{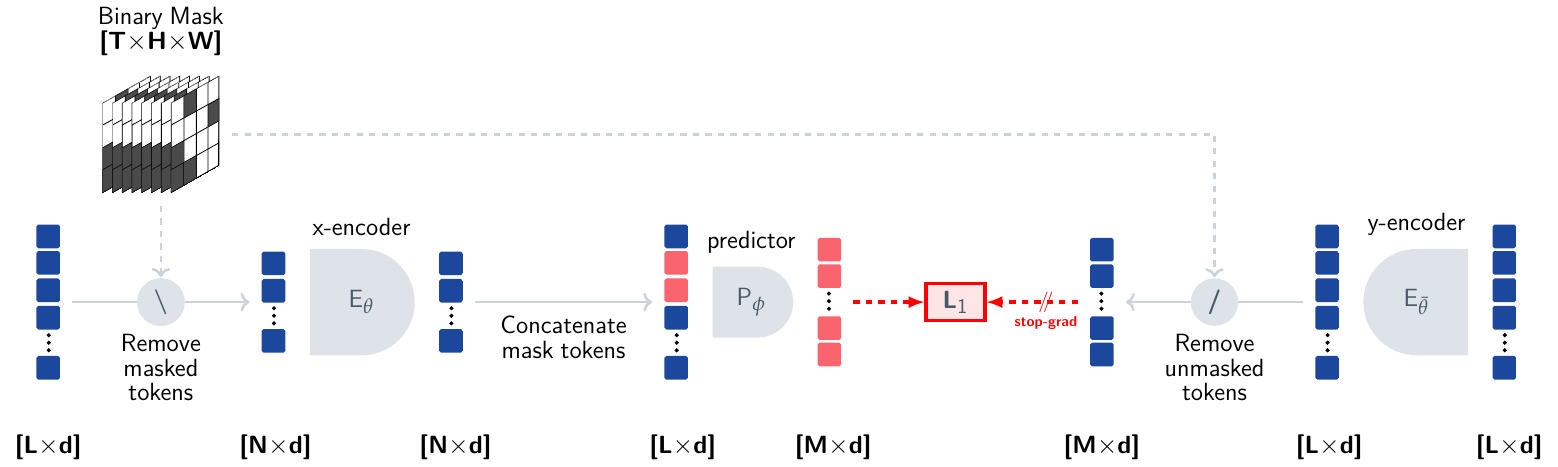}
    \caption{{\it \putalg.} Training operates on a video clip of $T$ frames with spatial resolution $H\times W$, flattened into a sequence of $L$ tokens.
    (Left to right):
    We first obtain the input of the $x$-encoder by dropping tokens from the video clip.
    The $x$-encoder then processes the masked video sequence, and outputs an embedding vector for each input token.
    Next, the outputs of the $x$-encoder are concatenated with a set of learnable mask tokens containing positional embeddings of the masked spatio-temporal patches.
    The predictor network processes the combined token sequence, and outputs an embedding vector for each mask token.
    The outputs of the predictor are then regressed to the prediction targets using an $L_1$ loss.
    The prediction targets correspond to the output of the $y$-encoder.}
    \label{fig:vjepa-complex}
\end{figure*}

\subsection{Prediction Task: Predicting $y$ from $x$}
\label{subsec:prediction_task}

The feature prediction task is based on a masked modeling formulation~\citep{he2021masked, tong2022videomae}; i.e., regions $x$ and $y$ from the video are sampled using masking.
To sample $y$ from a video, we sample several (possibly overlapping) spatially continuous blocks with various aspect ratios and repeat the spatial blocks across the entire temporal dimension of the video; $x$ is taken to be the complement.
Masking a large continuous block that covers the full temporal dimension limits information leakage due to the spatial and temporal redundancy of videos, and results in a harder prediction task~\citep{tong2022videomae}.

We leverage two types of masks: short-range masks, where we take the union of $8$ randomly sampled target blocks covering 15\% of each frame, and long-range masks, where we take the union of $2$ randomly sampled target blocks covering 70\% of each frame.
In both cases, the aspect ratio for all sampled blocks is randomly chosen in the range $(0.75, 1.5)$.
Given that both short-range and long-range masks are produced by sampling many blocks and taking their union, the result is an average masking ratio of $\sim90\%$.
We refer to our masking strategy as multi-block, and compare it to other possible masking strategies in Section~\ref{sec:ablations}.

\subsection{Network Parameterization}
We use a Vision Transformer (ViT)~\citep{dosovitskiy2020image,arnab2021vivit} as our video backbone.
To process a video with a transformer network, we split the video clip into a 3D grid of $L$ spatio-temporal patches, where a patch consists of a $16\times16$ pixel block spanning $2$ consecutive frames; we refer to these spatio-temporal patches as tokens.
This sequence of tokens is then directly processed by the stack of transformer blocks.
Inputs $x$ and $y$ correspond to masked regions of a video, we apply the video masks by simply dropping a subset of the tokens.
We apply masking at the input of the $x$-encoder, and at the output of the $y$-encoder to construct contextualized targets~\citep{baevski2022data2vec}.
The encoder is parameterized using standard ViT networks, while the predictor is a narrow transformer implemented using $12$ blocks with an embedding dimension of $384$.
Taking inspiration from masked autoencoders~\citep{he2021masked}, our predictor takes as input the sequence of embeddings produced by the $x$-encoder as well as a sequence of learnable mask tokens with positional embeddings indicating the spatio-temporal positions of the $y$ tokens.
The output of the predictor is an embedding vector for each mask token; see Figure~\ref{fig:vjepa-complex} and refer to Appendix~\ref{appendix:vjepa_extended_description} for more details.
\begin{table*}[t]
    \centering
    \caption{{\it Pixels vs.~Featurized Targets.} We ablate the effect of computing the prediction loss in feature space vs pixel space.  All models are trained on VideoMix2M for 90K iterations with a batch size of 3072 using the multi-block prediction task.
    We examine downstream performance using a frozen backbone with attentive probing, and report top-1 accuracy using a single center view. We also examine end-to-end fine-tuning performance of the models on K400. Predicting in feature space provide a consistent improvement over pixel space prediction.}
    \label{tb:pixels_vs_features}
    {\fontfamily{ptm}
    \fontsize{8.5pt}{8.5pt}\selectfont
    \begin{tabular}{cc ccc c}
    \toprule
    & & \multicolumn{3}{c}{\it Frozen Evaluation} & \multicolumn{1}{c}{\it Fine-Tuning} \\
    \cmidrule(l){3-5} \cmidrule(l){6-6}
     &  & \bf K400 & \bf SSv2 & \bf IN1K & \bf K400-ft\\
    \bf Target & \bf Arch.  &  {\fontsize{5.5pt}{5.5pt}\selectfont(16$\times$1$\times$1)} & {\fontsize{5.5pt}{5.5pt}\selectfont(16$\times$1$\times$1)} & & {\fontsize{5.5pt}{5.5pt}\selectfont(16$\times$5$\times$3)}\\
    \midrule
    Pixels & ViT-L/16 & 68.6 & 66.0 &  73.3 & 85.4\\
    Features & ViT-L/16 & \cc\bf 73.7 & \cc\bf 66.2 & \cc\bf 74.8 & \cc\bf 85.6\\
    \bottomrule
    \end{tabular}}
    \vskip 2em
    \centering
    {\fontfamily{ptm}
    \fontsize{8.5pt}{8.5pt}\selectfont
    \caption{{\it Pretraining Data Distribution.} We pretrain all models for 90K iterations using a batch size of 3072, and evaluate downstream performance of the frozen backbones with an attentive probe using a single center view. Average performance across tasks increases with the pretraining dataset size.}
    \label{tb:dataset_scale}
    \begin{tabular}{llr ccc c }
        \toprule
        & & & \multicolumn{3}{c}{\it Frozen Evaluation} & \\
        \cmidrule{4-6}
        & & & \bf K400 & \bf SSv2 & \bf IN1K & \bf Avg. \\
        \bf Arch. & \bf Data & \bf \#Samples  &  {\fontsize{4.5pt}{4.5pt}\selectfont(16$\times$1$\times$1)} & {\fontsize{4.5pt}{4.5pt}\selectfont(16$\times$1$\times$1)} & &  \\
        \midrule
        \multirow{4}{*}{ViT-L/16} & K710 & 700K & \bf 75.8 & 63.2 & 73.7 & 70.9 \\
        & K710+SSv2 & 900K & 72.9 & \bf 67.4 & 72.8 & 71.0 \\
        & K710+HT & 1900K & 74.5 & 64.2 & \bf 74.8 &  71.1 \\ \vspace{1ex}
        & VideoMix2M & 2000K & \cc 73.7 & \cc 66.2 & \cc \bf 74.8 & \cc \bf 71.5\\
        \multirow{2}{*}{ViT-H/16} & K710+SSv2 & 900K & \bf 75.7 & 66.8 & 73.7 & 72.0 \\
        & VideoMix2M & 2000K & \cc 74.0 & \cc \bf 68.5 & \cc \bf 75.9 & \cc \bf 72.8\\
        \bottomrule
    \end{tabular}}
\end{table*}

\subsection{Pretraining Data and Evaluation Setup}
\label{sec:pretraining_data_distribution}

\paragraph{\bf Pretraining.}
We combine several public datasets to construct an unsupervised video pretraining dataset, which we refer to as VideoMix2M.
Specifically, we combine the videos from HowTo100M (HT)~\citep{miech2019howto100m}, Kinetics-400/600/700 (K710)~\citep{kay2017kinetics}, and Something-Something-v2 (SSv2)~\citep{goyal2017something}, and remove any overlap with the validation sets of Kinetics-400/600/700 and Something-Something-v2, resulting in approximately 2 million videos.
We train a ViT-L/16, a ViT-H/16, and a ViT-H/16$_{384}$  transformer model on VideoMix2M.
We use a batch size of 3072 for the ViT-L/16 and ViT-H/16 models, and a batch size of 2400 for the ViT-H/16$_{384}$ model.
Each model takes as input a video clip of 16 frames sampled with a frame-skip of 4, corresponding to roughly 3 second clips on average.
The ViT-L/16 and ViT-H/16 process the video at a spatial resolution of 224, while the ViT-H/16$_{384}$ uses an input resolution of 384; cf.~Appendix~\ref{app:pretraining}.

\paragraph{\bf Evaluations.}
Pretrained models are evaluated on downstream video and image tasks.
On video tasks, we use a subset of the VideoGLUE benchmark~\citep{yuan2023videoglue} to test for various capabilities; specifically, we investigate action recognition on Kinetics-400 (K400)~\citep{kay2017kinetics}, motion classification on Something-Something-v2 (SSv2)~\citep{goyal2017something}, and action localization on AVA~\citep{gu2018ava}.
Action classification on Kinetics evaluates the appearance-based understanding of the model, as many action classes in the dataset can be inferred from the presence of specific objects in the video~\citep{sevilla2021only}.
Motion classification on Something-Something-v2 evaluates the temporal understanding of the model, as action classes in the dataset are decoupled from the appearance/presence of specific objects in the video~\citep{goyal2017something}.
Finally, action localization on AVA evaluates the ability of the model to understand and localize motions in the video.
We follow standard practice and report accuracy on K400 and SSv2 by sampling several spatial and temporal views.
For static image tasks, we explore object recognition on ImageNet~\citep{russakovsky2015imagenet}, scene classification on Places205~\citep{places205}, and fine-grained recognition on iNaturalist 2021~\citep{van2018inaturalist}.

\section{What Matters for Learning Representations from Video?}
\label{sec:ablations}

In this section we isolate the contributions of several design choices, including: a) the use of a feature prediction versus pixel prediction objective, b) the construction of the pretraining data distribution, c) the feature pooling strategy for leveraging the model's representations in downstream tasks, and d) the masking strategy, towards identifying: what to predict from what?

\subsection{Predicting Representations versus Pixels}
\label{subsec:pixel_ablation}
We first ablate the effect of computing the prediction loss in representation space.
We train a pair of ViT-L/16 models using either a \putalg feature prediction loss, or a mean-squared error loss with the normalized pixel values, as in masked autoencoders~\citep{he2021masked}, and perform a sweep over the learning rate and weight decay schedules for both approaches.
All models are pretrained on VideoMix2M for 90K iterations with a batch size of 3072 using multi-block masking.
We examine performance on Kinetics-400 (K400), Something-Something-v2 (SSv2), and ImageNet-1K (IN1K), using a frozen backbone with an attentive probe, and report top-1 accuracy using a single center view.
We also examine end-to-end fine-tuning performance of the models on Kinetics-400.

Results of this comparison are reported in Table~\ref{tb:pixels_vs_features} and indicate that predicting in feature space provides a consistent performance improvement over pixel space prediction in both frozen evaluation of the video backbone, as well as end-to-end fine-tuning.

\subsection{Pretraining Data Distribution}
Next we study the impact of the pretraining data distribution in Table~\ref{tb:dataset_scale}.
Leveraging large scale datasets has been critical for enabling the surge of advancements in other modalities, such as text and images~\citep{kaplan2020scaling, cherti2023reproducible}.
We investigate whether a similar trend holds for video data.
To control for the possible confounding variable of compute budget, we pretrain all models in Table~\ref{tb:dataset_scale} for 90K iterations using a batch-size of 3072.
We report downstream results on K400, SSv2, and IN1K using a frozen backbone with an attentive probe, and report top-1 accuracy using a single center view.

Table~\ref{tb:dataset_scale} shows that average performance across tasks monotonically increases as we increase the size of the pretraining dataset, but the best task-specific performance is obtained by independently selecting the pretraining data for each specific downstream task.
For instance, the L/16 obtains its best SSv2 performance when pretrained on K710+SSv2, its best K400 performance when pretrained only on K710, and its best IN1K performance when pretrained only on K710+HT. The best average performance across all tasks is achieved by pretraining  VideoMix2M, which combines all the data sources.
Similarly, the H/16 pretrained on K710+SSv2 achieves a greater K400 score than the H/16 pretrained on VideoMix2M, however, the top performing H/16 on average is pretrained on VideoMix2M.
\begin{table}[t]
    \centering
    {\fontfamily{ptm}
    \fontsize{8.5pt}{8.5pt}\selectfont
    \caption{{\it Average Pooling vs. Adaptive Pooling.} 
    We pool the feature map output by the frozen \putalg encoder using an attentive probe, which is then fed into a linear classifier for downstream supervised tasks (K400 and SSv2).
    We evaluate two pooling strategies: 1) average pooling (Avg.), and attentive pooling (Att.).
    Results are reported using a single center view.
    Using adaptive pooling with a cross-attention layer leads to improvements of $+17.3$ points on K400 and $+16.1$ points on SSv2.}
    \label{tb:adaptive_feature_pooling}
    \begin{tabular}{ll cccc}
        \toprule
        & & \multicolumn{4}{c}{\it Frozen Evaluation} \\
        \cmidrule{3-6}
        & & \multicolumn{2}{c}{\bf K400} & \multicolumn{2}{c}{\bf SSv2}\\
        & & \multicolumn{2}{c}{{\fontsize{4.5pt}{4.5pt}\selectfont(16$\times$1$\times$1)}} & \multicolumn{2}{c}{{\fontsize{4.5pt}{4.5pt}\selectfont(16$\times$1$\times$1)}}\\
        \bf Method & \bf Arch. & Avg. & Att. & Avg. & Att.\\
        \midrule
        \putalg & ViT-L/16 & \cc 56.7 & \cc \bf 73.7 & \cc 50.1 & \cc \bf 66.2\\
        \bottomrule
    \end{tabular}}
\end{table}

\subsection{Evaluation: Attentive Probing}
Next we explore the feature pooling strategy for applying the model's representations in downstream tasks.
Since the prediction objective in equation~\eqref{eq:loss} is unnormalized, there is no a priori reason for the encoder to yield a linearly separable subspace~\citep{chen2020simple}.
Thus, rather than using a linear operation (averaging) to pool the features output of the frozen backbone, we explore a learnable non-linear pooling strategy. 
Specifically, when evaluating the frozen pretrained backbone on downstream tasks, we learn a cross-attention layer with a learnable query token.
The output of the cross-attention layer is then added back to the query token (residual connection), and then fed into two-layer MLP with a single GeLU activation, followed by a LayerNorm, and finally a linear classifier.

In Table~\ref{tb:adaptive_feature_pooling} we see that using adaptive pooling with a  learnable cross-attention layer leads to a significant improvement of $+17$ points on K400 and $+16.1$ points on SSv2.
Using an attentive-probe is also beneficial for other baseline models as reported in Appendix~\ref{app:extra_results}.

\subsection{Prediction Task: Predicting $y$ from $x$}
We conduct an ablation on the masking strategy used in \putalg pretraining.
We examine the following masking strategies: {\tt random-tube[r]} in which $x$ is obtained by removing a random fraction $r$ of tubes (spatial patches extended across the entire temporal duration) from the video, {\tt causal multi-block[p]} in which $x$ is restricted to the first $p$ frames of the 16-frame video, which are then masked with a random set of spatio-temporal blocks, and {\tt multi-block} in which $x$ obtained by masking a random set of spatio-temporal blocks from the entire video.
Spatio-temporal blocks are sampled using the parameters described in Section~\ref{subsec:prediction_task}; an ablation on the size and quantity of masked spatio-temporal blocks is provided in Appendix~\ref{app:masking_ablation}.
\begin{table}[t]
    \centering
    {\fontfamily{ptm}
    \fontsize{8.5pt}{8.5pt}\selectfont
    \caption{{\it Ablating Prediction Task.} Models are ViT-L/16 networks pretrained on K710 and SSv2 and evaluated with an attentive probe using a single center view. The region $x$ is sampled by masking spatio-temporal regions in the video; $y$ is the mask complement.
    {\bf 1) random-tube[r]:} $x$ is obtained by masking a fraction $r$ of tubes (spatial patches extended across the entire temporal duration) from the video, {\bf 2) causal multi-block[p]:} $x$ is restricted to the first $p$ frames of the 16-frame video, which are then masked with a random set of spatio-temporal blocks, {\bf 3) multi-block}: $x$ is obtained by masking a random set of spatio-temporal blocks from the entire video. Best performance obtained by using multiblock masking.}
    \label{tb:masking}
    \begin{tabular}{l c c c}
        \toprule
        & \multicolumn{3}{c}{\it Frozen Evaluation} \\
        \cmidrule{2-4}
        & \bf K400 & \bf SSv2 & \bf IN1K \\
        \bf Masking & {\fontsize{5.5pt}{5.5pt}\selectfont(16$\times$1$\times$1)} & {\fontsize{5.5pt}{5.5pt}\selectfont(16$\times$1$\times$1)} & \\
        \midrule
        \tt random-tube[0.9] & 51.5 & 46.4 & 55.6 \\
        \tt causal multi-block[6] & 61.3 & 49.8 & 66.9 \\
        \tt causal multi-block[12] & 71.9 & 63.6 & 72.2 \\
        \tt multi-block & \cc\bf 72.9 & \cc\bf 67.4 & \cc\bf 72.8\\
        \bottomrule
    \end{tabular}}
\end{table}

Table~\ref{tb:masking} indicates that the best results are obtained by sampling $x$ using a {\it multi-block} strategy, wherein the network is forced to make predictions after removing large continuous blocks in the video.
When $x$ is only sampled from the first few frames of the video, as in the {\it causal multi-block} strategy, we observe a decrease in downstream performances.
Finally, the {\it random-tube} strategy, wherein  90\% of the tubes in the video are randomly masked, leads to features of low-semantic quality when combined with our feature prediction objective. 

\begin{table*}[t]
    \centering
    {\fontfamily{ptm}\fontsize{7pt}{7pt}\selectfont
    \caption{{\bf \it Comparison with Pixel Prediction Methods.} We compare \putalg with OmniMAE~\citep{girdhar2023omnimae}, VideoMAE~\citep{tong2022videomae}, and Hiera~\citep{ryali2023hiera}, which leverage a pixel-reconstruction loss.
    All models are trained using a ViT-L architecture or a comparable Hiera-L.
    We evaluate the approaches on downstream image tasks (IN1K, Places205, iNat201) and video tasks (K400, SSv2, AVA) in both frozen evaluation (with a frozen backbone), and end-to-end fine-tuning.
    All models are evaluated at resolution 224.
    On K400 and SSv2 we follow the standard practice of reporting accuracy from several spatial and temporal views from the video. 
    In frozen evaluation, \putalg outperforms the baselines on all downstream tasks, except ImageNet, where the model achieves $74.8\%$ compared to $75.1\%$ of an OmniMAE model trained directly on ImageNet.
    \putalg also achieves the best fine-tuning performance amongs all ViT-L models and matches the Hiera-L on SSv2.
    The \putalg results are achieved while processing significantly fewer examples during pretraining.
    }
    \label{tb:pixel_comparison}
    \begin{tabular}{llrr cccccc cc}
    \toprule
    & & & & \multicolumn{6}{c}{ \it Frozen Evaluation w/ Att.~Pooling} & \multicolumn{2}{c}{\it Fine-Tuning} \\
    \cmidrule(l){5-10} \cmidrule(l){11-12}
    & & \bf \#Samples & & \bf K400 & \bf SSv2 & \bf AVA & \bf IN1K & \bf Places205 & \bf iNat21 & \bf K400-ft & \bf SSv2-ft \\
    \bf Method & \bf Arch. & \bf Seen & \bf Iter. & {\fontsize{5.5pt}{5.5pt}\selectfont(16$\times$8$\times$3)} & {\fontsize{5.5pt}{5.5pt}\selectfont(16$\times$2$\times$3)} & & & & & {\fontsize{5.5pt}{5.5pt}\selectfont(16$\times$5$\times$3)} & {\fontsize{5.5pt}{5.5pt}\selectfont(16$\times$2$\times$3)}\\
    \midrule
    \multicolumn{5}{l}{\bf\it Methods pretrained using pixel prediction} & & & & & \\[1ex]
    OmniMAE & ViT-L/16 & 2400M & 1170K & 65.6 & 60.6 & 14.4 & \bf 75.1 & 59.8 & 66.1 & 84.0 & 74.2 \\
    VideoMAE & ViT-L/16 & 410M & 400K & 77.8 & 65.5 & 21.6 & 71.1 & 59.3 & 64.6 & 85.4 & 74.3 \\
    Hiera & Hiera-L & 770M & 1500K & 75.5 & 64.2 & 15.8 & 68.9 & 58.5 & 56.9 & \bf 87.3 & \bf 75.1 \\
    \midrule
    V-JEPA & ViT-L/16 & 270M & 90K & \cc\bf 80.8 & \cc\bf 69.5 & \cc\bf 25.6 & \cc 74.8 & \cc\bf 60.3 & \cc\bf 67.8 & \cc 85.6 & \cc \bf 75.1 \\
    \bottomrule
    \end{tabular}}
    \vskip 1em
    \centering
    {\fontfamily{ptm}\fontsize{7pt}{7pt}\selectfont
    \caption{{\it Comparison with State-of-the-Art Models.}
    We compare \putalg with state-of-the-art baselines in frozen evaluation with an attentive probe on downstream image tasks (IN1K, Place205, iNat21) and video tasks (K400, SSv2, AVA).
    All models are evaluated at resolution 224, except I-JEPA$_{512}$ and V-JEPA$_{384}$ which are evaluated respectively at resolution $512$ and $384$.
    On K400 and SSv2 we follow the standard practice of reporting accuracy from several spatial and temporal views from the video. 
    Compared to other video baselines, \putalg exhibits a consistent improvement across all downstream tasks.
    Compared to image-models that excel under the frozen evaluation, \putalg shows a significant performance improvement on tasks requiring motion understanding (+21 points on SSv2), and reduces the gap between video and image models on tasks requiring static appearance-based features.}
    \label{tb:large_results}\vspace{1em}
    \begin{tabular}{llrr ccc ccc}
        \toprule
        & & & & \multicolumn{3}{c}{\it Video Tasks} & \multicolumn{3}{c}{\it Image Tasks} \\
        \cmidrule(l){5-7} \cmidrule(l){8-10}
        & & & & \bf K400 & \bf SSv2 & \bf AVA & \bf IN1K & \bf Places205 & \bf iNat21 \\
        \bf Method & \bf Arch. & \bf Params. & \bf Data & {\fontsize{5.5pt}{5.5pt}\selectfont(16$\times$8$\times$3)} & {\fontsize{5.5pt}{5.5pt}\selectfont(16$\times$2$\times$3)} & & & & \\
        \midrule
        \multicolumn{2}{l}{\bf\it Methods pretrained on Images} & & & & & & \\[1ex]
        I-JEPA & ViT-H/16$_{512}$ & 630M & IN22K & 79.7 & 50.0 & 19.8 & 84.4 & 66.5 & 85.7 \\
        OpenCLIP & ViT-G/14 & 1800M & LAION & 81.8 & 34.8 & 23.2 & 85.3 & \bf 70.2 & 83.6 \\
        DINOv2 & ViT-g/14 & 1100M & LVD-142M & \bf 83.4 & 50.6 & 24.3 & \bf 86.2 & 68.4 & \bf 88.8 \\
        \midrule
        \multicolumn{2}{l}{\bf\it Methods pretrained on Videos} & & & & & & \\[1ex]
        MVD & ViT-L/16 & 200M & IN1K+K400 & 79.4 & 66.5 & 19.7 & 73.3 & 59.4 & 65.7 \\
        OmniMAE & ViT-H/16 & 630M & IN1K+SSv2 & 71.4 & 65.4 & 16.0 & 76.3 & 60.6 & 72.4\\
        VideoMAE & ViT-H/16 & 630M & K400 & 79.8 & 66.2 & 20.7 & 72.3 & 59.1 & 65.5 \\
        VideoMAEv2 & ViT-g/14 & 1100M & Un.Hybrid & 71.2 & 61.2 & 12.9 & 71.4 & 60.6 & 68.3\\
        Hiera & Hiera-H & 670M & K400 & 77.0 & 64.7 & 17.5 & 71.4 & 59.5 & 61.7 \\
        \midrule
        \multirow{3}{*}{V-JEPA} & ViT-L/16 & 200M & \multirow{3}{*}{VideoMix2M}
        & \cc 80.8 & \cc 69.5 & \cc 25.6 & \cc 74.8 & \cc 60.3 & \cc 67.8\\
        & ViT-H/16 & 630M & & \cc\bf 82.0 & \cc 71.4 & \cc\bf 25.8 & \cc 75.9 & \cc 61.7 & \cc 67.9 \\
        & ViT-H/16$_{384}$ & 630M & & \cc 81.9 & \cc\bf 72.2 & \cc 25.0 & \cc\bf 77.4 & \cc\bf 62.8 & \cc \bf 72.6 \\
        \bottomrule
    \end{tabular}}
\end{table*}

\section{Comparison with Prior Work}
In Section~\ref{subsec:pixel_comparison}, we investigate the impact of feature prediction by comparing \putalg with video approaches that rely on pixel prediction, while using a similar architecture for all baselines.
Subsequently, in Section~\ref{subsec:sota_comparison}, we remove the architectural constraint and report the best performance across architectures for self-supervised video and image pretraining approaches. 
Finally, we explore the label-efficiency of \putalg relative to other self-supervised video pretraining approaches in Section~\ref{subsec:lowshot}.
We further detail the evaluation setup in Appendix~\ref{app:evaluation}.

\subsection{Comparison with Pixel Prediction}
\label{subsec:pixel_comparison}

To investigate the effectiveness of feature prediction pretraining, we first compare \putalg to video masked modeling models relying on a pixel prediction loss.
We control for the possible confounding factor of model architecture by evaluating all models using either a ViT-L/16 encoder, or a Hiera-L encoder, which has a similar number of parameters.
For the pixel prediction baselines we consider VideoMAE~\citep{tong2022videomae, wang2023videomae}, which trains vision transformer autoencoders exclusively on video, Hiera~\citep{ryali2023hiera}, which trains a hierarchical transformer autoencoder on video, and OmniMAE~\citep{girdhar2023omnimae}, which trains a vision transformer autoencoder on static images and video simultaneously.

Table~\ref{tb:pixel_comparison} examines both frozen evaluation with an attentive probe on downstream video and image tasks, as well as end-to-end fine-tuning.
In frozen evaluation, \putalg outperforms the baselines on all downstream tasks, except ImageNet, where we achieve $74.8\%$ compared to $75.1\%$ of an OmniMAE model trained directly on ImageNet; hence, \putalg achieves comparable ImageNet performance despite only pretraining on video.

\begin{figure}[t]
    \includegraphics[width=\linewidth]{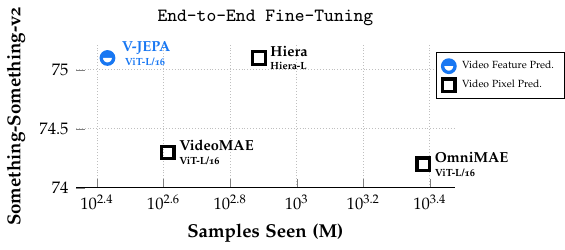}
    \caption{{\it SSv2 fine-tuning performance vs.~Samples Seen.} We report SSv2 fine-tuning for \putalg and pixel-reconstruction baselines using a ViT-L/16 or Hiera-L architecture. \putalg outperforms all pixel-reconstruction methods using a ViT-L/16 and matches the Hiera-L performance while seeing significantly less samples during pretraining.}
    \label{fig:ssv2_finetuning}
\end{figure}

Under the fine-tuning protocol, \putalg also achieves the best performance of any model trained with a ViT-L/16, and matches the performance of the Hiera-L on SSv2, which benefits from a hierachical prior~\citep{ryali2023hiera}.
The \putalg models achieve this result while processing significantly fewer samples during pretraining (Figure~\ref{fig:ssv2_finetuning}), demonstrating the efficiency of feature prediction as a learning principle.

\subsection{Comparison with State-of-the-Art}
\label{subsec:sota_comparison}

Next, in Table~\ref{tb:large_results}, we inspect how the \putalg models pretrained on video stack up next to the largest state-of-the-art self-supervised image and video models when freezing the backbone encoder and training an attentive probe on top.
Our image pretrained baselines include OpenCLIP~\citep{cherti2023reproducible}, DINOv2~\citep{oquab2023dinov2}, and I-JEPA~\citep{assran2023self}.
The OpenCLIP model is trained with a contrastive image-text alignment objective, DINOv2 and I-JEPA are trained with self-supervision.
These models are known to excel in their frozen-evaluation performance~\citep{oquab2023dinov2}; i.e., their ability to produce visual features that can be applied to many downstream tasks simultaneously, without end-to-end fine-tuning, and thus provide highly competitive baselines.
Our video pretrained baselines include VideoMAE~\citep{tong2022videomae}, OmniMAE~\citep{girdhar2023omnimae}, Hiera~\citep{ryali2023hiera}, VideoMAEv2~\citep{wang2023videomae}, and MVD~\citep{wang2023masked}.
The OpenCLIP, DINOv2 and VideoMAEv2 models are parameterized as Giant/Gigantic vision transformer architectures containing over 1B parameters trained on large-scale image or video datasets.
\begin{figure}[t]
    \includegraphics[width=\linewidth]{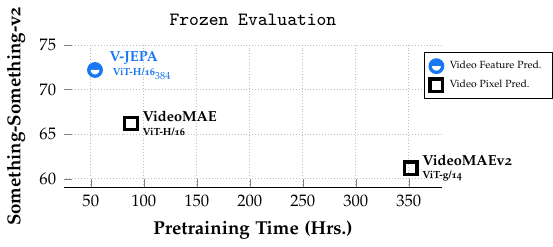}
    \caption{{\it SSv2 frozen-evaluation performance vs.~Pretraining Time.} 
    Wallclock times for all methods are measured on a single GPU with a batch size of 10 clips, using the official codebases for VideoMAE and VideoMAEv2, and linearly extrapolated assuming a global batch size of 2400 samples.
    However, note that the SSv2 accuracies of video pixel prediction methods are actually obtained with small batch sizes and significantly longer training schedules.
    \putalg outperforms pixel-reconstruction methods while training significantly faster.}
    \label{fig:ssv2_frozen}
\end{figure}
\begin{table*}[t]
    \centering
    {\fontfamily{ptm}\fontsize{7pt}{7pt}\selectfont
    \caption{{\it Low-Shot Frozen Evaluation.} 
        Comparing \putalg to other video models in frozen evaluation on Kinetics-400 and Something-Something-v2 as we vary the percentage of labeled examples from each dataset available for training the attentive probe.
        We train the probes in several low-shot settings: using either 5\% of the train set, 10\%, or 50\%, and take 3 random splits in each setting to obtain more robust metrics, resulting in 9 different evaluation experiments for each model.
        We report the mean performances and standard deviation using the K400 and SSv2 validation sets.
        \putalg is more label-efficient than other models; specifically, decreasing the available number of labeled examples from each class increases the performance gap between \putalg and the baselines.}
    \label{tb:lowshot}
    \begin{tabular}{ll ccc ccc}
        \toprule
        & & \multicolumn{6}{c}{\it Frozen Evaluation} \\[1ex]
        & & \multicolumn{3}{c}{\bf K400} & \multicolumn{3}{c}{\bf SSv2} \\
        & & \multicolumn{3}{c}{\fontsize{5.5pt}{5.5pt}\selectfont(16$\times$8$\times$3)} & \multicolumn{3}{c}{\fontsize{5.5pt}{5.5pt}\selectfont(16$\times$2$\times$3)} \\
        \cmidrule(l){3-5} \cmidrule(l){6-8}
        & & 5\% & 10\% & 50\% & 5\% & 10\% & 50\% \\
        \bf Method & \bf Arch. & \fontsize{5.5pt}{5.5pt}\selectfont($\sim$29 samples per class) & \fontsize{5.5pt}{5.5pt}\selectfont($\sim$58 samples per class) & \fontsize{5.5pt}{5.5pt}\selectfont($\sim$287 samples per class) & \fontsize{5.5pt}{5.5pt}\selectfont($\sim$48 samples per class) & \fontsize{5.5pt}{5.5pt}\selectfont($\sim$96 samples per class) & \fontsize{5.5pt}{5.5pt}\selectfont($\sim$440 samples per class) \\
        \midrule
        MVD & ViT-L/16 & 62.6 $\pm$ 0.2 & 68.3 $\pm$ 0.2 & 77.2 $\pm$ 0.3 & 42.9 $\pm$ 0.8 & 49.5 $\pm$ 0.6 & 61.0 $\pm$ 0.2 \\
        VideoMAE & ViT-H/16 & 62.3 $\pm$ 0.3 & 68.5 $\pm$ 0.2 & 78.2 $\pm$ 0.1 & 41.4 $\pm$ 0.8 & 48.1 $\pm$ 0.2 & 60.5 $\pm$ 0.4 \\
        VideoMAEv2 & ViT-g/14 & 37.0 $\pm$ 0.3 & 48.8 $\pm$ 0.4 & 67.8 $\pm$ 0.1 & 28.0 $\pm$ 1.0 & 37.3 $\pm$ 0.3 & 54.0 $\pm$ 0.3 \\
        \midrule
        \multirow{2}{*}{V-JEPA} & ViT-H/16 & \cc 67.0 $\pm$ 0.2 & \cc 72.1 $\pm$ 0.1 & \cc 80.2 $\pm$ 0.2 & \cc 51.9 $\pm$ 0.3 & \cc 57.5 $\pm$ 0.4 & \cc 67.3 $\pm$ 0.2 \\
        & ViT-H/16$_{384}$ & \bf\cc 68.2 $\pm$ 0.2 & \cc\bf 72.8 $\pm$ 0.2 & \bf\cc 80.6 $\pm$ 0.2 &\bf\cc  54.0 $\pm$ 0.2 & \bf\cc 59.3 $\pm$ 0.5 & \bf\cc 67.9 $\pm$ 0.2 \\
        \bottomrule
    \end{tabular}}
\end{table*}

\paragraph{\bf Comparison with video models.}
Compared to large-scale video baselines, the \putalg models outperform all previous models on every downstream video and image task with notable margin (see Table~\ref{tb:large_results}).
Our H/16 model outperforms the largest publicly available VideoMAE, VideoMAEv2, OmniMAE, MVD, and Hiera models by at least $+5$ points in motion understanding (Something-Something-v2), $+2$ points in action recognition (Kinetics-400), $+5$ points on action detection (AVA), $+1$ point on object recognition (ImageNet-1K), $+2$ points in scene recognition (Places205), and $+0.2$ points on fine-grained recognition (iNaturalist).
Moreover, when comparing pretraining wallclock time in Figure~\ref{fig:ssv2_frozen}, we see that \putalg achieves this performance with a roughly $2\times$ speedup compared to the large pixel prediction models.

\paragraph{\bf Comparison with image models.}
On tasks that require a fine-grained understanding of motion (Something-Something-v2), the \putalg models provide a major improvement (over $+21$ points) compared to large-scale image baselines, such as DINOv2, OpenCLIP, and I-JEPA.
Self-supervised pretraining from videos allows to model dynamic concepts that are not easily learned from static image datasets.
Similarly, we observe that the \putalg models outperform image-based pretraining on action localization.

On Kinetics-400, we find image models to perform well; e.g., while DINOv2~\citep{oquab2023dinov2} previously reported $78.4\%$ on K400 with a linear probe, we improve the frozen evaluation of the g/14 model to $83.4\%$ by using an attentive probe.
In this case, our H/16 model achieves $82.0\%$ top-1 accuracy.
It is worth noting that the label for many Kinetics videos can be inferred using appearance-based cues, without requiring an understanding of motion~\citep{sevilla2021only}.

The \putalg models narrow the gap with image models on image classification tasks.
In particular, \putalg achieves a score of $77.4\%$ on ImageNet using a one-layer attentive probe, which can be further improved to $\bf{77.9\%}$ using a two-layer attentive probe.
More generally, we hypothesize that the datasets used to train \putalg and other video models are too constrained and lack the visual diversity of the internet-scale pretraining data used by the images models; as such, there is value in focusing future work on building diverse publicly available video datasets.

\subsection{Label-efficiency}
\label{subsec:lowshot}
We examine the label-efficiency of \putalg compared to other self-supervised video models by measuring the ability of the pretrained backbones to adapt to downstream tasks with few labels.
Specifically, we investigate the performance of the frozen models on Kinetics-400 and Something-Something-v2 as we vary the percentage of labeled examples from each dataset available for training the attentive probe.
We train the probes in several low-shot settings: using either 5\% of the train set, 10\%, or 50\%, and take 3 random splits in each setting to obtain more robust metrics, resulting in 9 different evaluation experiments for each model.
Table~\ref{tb:lowshot} reports the mean performances and standard deviation using the  K400 and SSv2 validation sets.

We find \putalg to be more label-efficient than other self-supervised video models: decreasing the available number of labeled examples for training the attentive probe results in an increase in the performance gap between \putalg and the other models.
In particular, the performance of the largest \putalg model on K400 drops by 12\% to 68.2\% top-1 when we reduce the number of labeled examples by a factor of $10\times$ (from roughly 287 examples per class to 29 examples per class).
By contrast, VideoMAEv2 drops by 30\% to 37.0\% top-1, VideoMAE drops by 15.9\% to 62.3\% top-1, and MVD drops by 14.6\% to 62.6\% top-1.

Similar observations hold on SSv2.
The performance of the largest \putalg model on SSv2 drops by 13.9\% to 54.0\% top-1 when we reduce the number of labeled examples by a factor of $10\times$ (from roughly 440 examples per class to 48 examples per class).
By contrast, VideoMAEv2 drops by 26\% to 28.0\% top-1, VideoMAE drops by 19.1\% to 41.4\% top-1, and MVD drops by 18.1\% to 42.9\% top-1.

\section{Evaluating the Predictor}
\begin{figure*}[t!]
    \centering
    \begin{subfigure}[b]{\textwidth}
        \centering
        \includegraphics[width=0.825\linewidth]{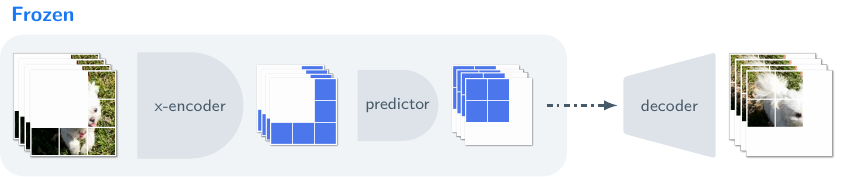}
        \caption{
        {\bf Visualization Methodology.}
        We train a conditional diffusion model to decode the \putalg feature-space predictions to  interpretable pixels; the pretrained \putalg encoder and predictor networks are kept frozen in this process.
        The decoder is only fed the representations predicted for the missing regions of the video, and does not have access to the unmasked regions of the video.
        }
        \label{fig:decoder_method}
    \end{subfigure}
    \vskip 4mm
    \begin{subfigure}[b]{\textwidth}
        \centering
        \includegraphics[width=0.485\linewidth]{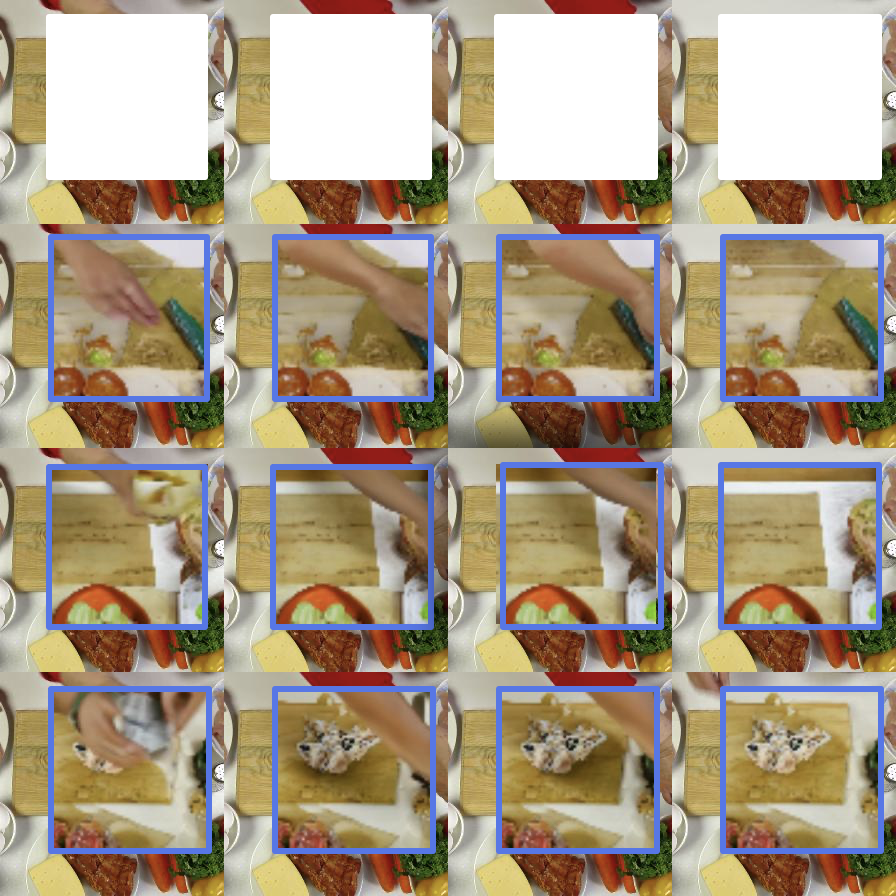}\quad
        \includegraphics[width=0.485\linewidth]{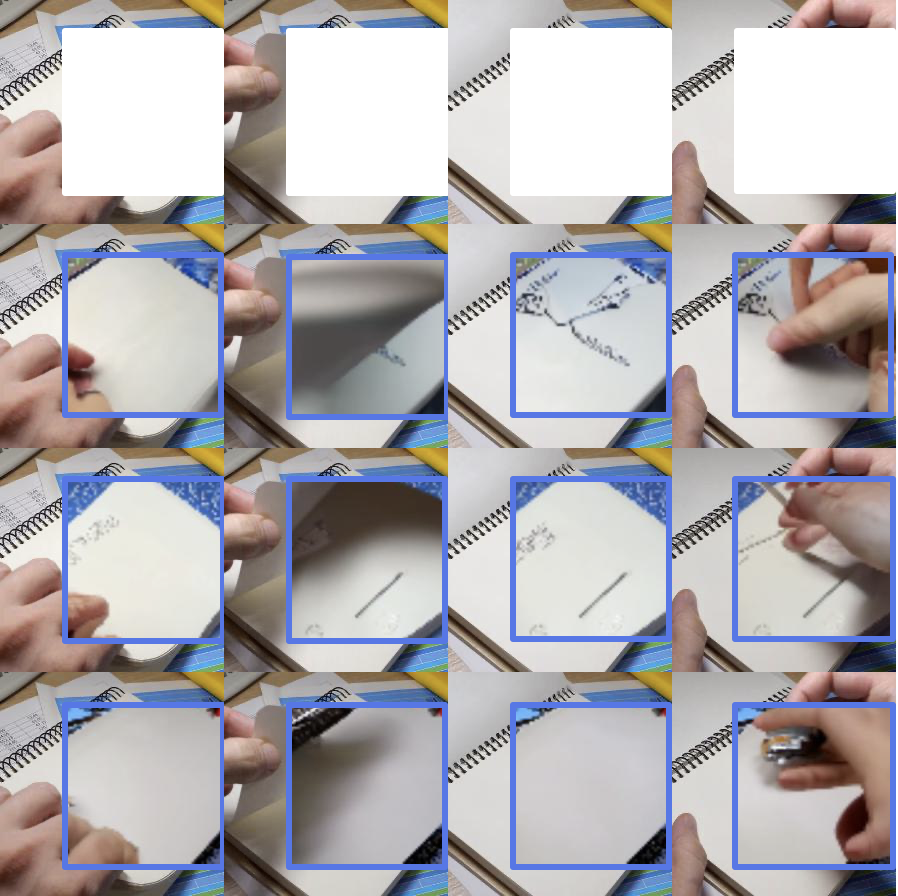}
        \caption{
        {\bf Visualizations.}
        {\it First Row:} Masked videos used as input to the \putalg models (a pretrained ViT-H/16 encoder and its corresponding predictor network). {\it Other rows:} Bounding boxes contain various samples from the decoder overlayed on the original video.
        \putalg is not a generative model and the decoder does not have access to the context (first row), so we do not expect samples to exactly match the input.
        This experiment qualitatively illustrates what information is encoded and predicted by \putalg.
        In particular, characteristics that are common across samples represent information that is encoded in the \putalg predictions.
        \putalg generates predictions that are spatially and temporally coherent with unmask region of the video. The predictions also capture consistent motion through time.
        }
        \label{fig:prediction-sample}
    \end{subfigure}
    \caption{{\it Qualitative Analysis.} Offline visualizations of the \putalg feature-space predictions.}
    \label{fig:prediction-visualization}
\end{figure*}

Next, we seek to qualitatively inspect the \putalg models.
Recall that the predictor network in \putalg predicts the representations of a masked spatio-temporal region $y$ from a visible region $x$, given the positional information of the masked regions (see Section~\ref{sec:methodology}).
To qualitatively investigate the grounding of the feature-space predictions, we freeze the pretrained encoder and predictor networks and train a conditional diffusion decoder to map the \putalg predictions to interpretable pixels.
Notably, the decoder is only fed the representations predicted for the missing regions of the video, and does not have access to the unmasked regions of the video (see Figure~\ref{fig:decoder_method}).

Given a masked video, we use the \putalg pretrained models to predict the representations of the missing regions, and then use the decoder to project the representations to pixel space. Figure~\ref{fig:prediction-sample} shows decoder outputs for various random seeds.
Qualities that are common across samples represent information that is contained in the predictor representation.

Figure~\ref{fig:prediction-sample} shows that the \putalg feature predictions are indeed grounded, and exhibit spatio-temporal consistency with the unmasked regions of the video.
Specifically, the samples in Figure~\ref{fig:prediction-sample} show that the \putalg predictor correctly captures positional uncertainty and produces a variety of visual objects at various locations with consistent motion.
Some of the samples also demonstrate an understanding of object-permanence, as the visual objects remain consistent after partial occlusion.

\section{Conclusion}
In this work, we explored the effectiveness of feature prediction as a stand-alone objective for unsupervised learning from video and introduced \putalg, a collection of vision models trained solely using a self-supervised feature prediction objective.
The \putalg models demonstrate the ability to solve various downstream image and video tasks without adaption of the model parameters, and outperform previous video representation learning approaches in frozen evaluation on action recognition, spatio-temporal action detection, and image classification tasks.
Additionally, we show that pretraining \putalg on videos is particularly effective for solving downstream tasks requiring fine-grained motion understanding, while large-scale image models trained on internet scale datasets fall short on such tasks.
Finally, we empirically observed that \putalg models are label-efficient learners, and exhibit good performance on downstream tasks, even when only few labeled examples are available.

\bibliographystyle{assets/plainnat}
\bibliography{paper}

\begin{thebibliography}{81}
\providecommand{\natexlab}[1]{#1}
\providecommand{\url}[1]{\texttt{#1}}
\expandafter\ifx\csname urlstyle\endcsname\relax
  \providecommand{\doi}[1]{doi: #1}\else
  \providecommand{\doi}{doi: \begingroup \urlstyle{rm}\Url}\fi

\bibitem[Akbari et~al.(2021)Akbari, Yuan, Qian, Chuang, Chang, Cui, and Gong]{akbari2021vatt}
Hassan Akbari, Liangzhe Yuan, Rui Qian, Wei-Hong Chuang, Shih-Fu Chang, Yin Cui, and Boqing Gong.
\newblock Vatt: Transformers for multimodal self-supervised learning from raw video, audio and text.
\newblock \emph{Advances in Neural Information Processing Systems}, 34:\penalty0 24206--24221, 2021.

\bibitem[Arnab et~al.(2021)Arnab, Dehghani, Heigold, Sun, Lucic, and Schmid]{arnab2021vivit}
Anurag Arnab, Mostafa Dehghani, Georg Heigold, Chen Sun, Mario Lucic, and Cordelia Schmid.
\newblock Vivit: A video vision transformer.
\newblock In \emph{Proceedings of the IEEE international conference on computer vision}, 2021.

\bibitem[Assran et~al.(2022)Assran, Caron, Misra, Bojanowski, Bordes, Vincent, Joulin, Rabbat, and Ballas]{assran2022masked}
Mahmoud Assran, Mathilde Caron, Ishan Misra, Piotr Bojanowski, Florian Bordes, Pascal Vincent, Armand Joulin, Michael Rabbat, and Nicolas Ballas.
\newblock Masked siamese networks for label-efficient learning.
\newblock \emph{arXiv preprint arXiv:2204.07141}, 2022.

\bibitem[Assran et~al.(2023)Assran, Duval, Misra, Bojanowski, Vincent, Rabbat, LeCun, and Ballas]{assran2023self}
Mahmoud Assran, Quentin Duval, Ishan Misra, Piotr Bojanowski, Pascal Vincent, Michael Rabbat, Yann LeCun, and Nicolas Ballas.
\newblock Self-supervised learning from images with a joint-embedding predictive architecture.
\newblock In \emph{Proceedings of the IEEE/CVF Conference on Computer Vision and Pattern Recognition}, pages 15619--15629, 2023.

\bibitem[Baevski et~al.(2022{\natexlab{a}})Baevski, Babu, Hsu, and Auli]{baevski2022efficient}
Alexei Baevski, Arun Babu, Wei-Ning Hsu, and Michael Auli.
\newblock Efficient self-supervised learning with contextualized target representations for vision, speech and language.
\newblock \emph{arXiv preprint arXiv:2212.07525}, 2022{\natexlab{a}}.

\bibitem[Baevski et~al.(2022{\natexlab{b}})Baevski, Hsu, Xu, Babu, Gu, and Auli]{baevski2022data2vec}
Alexei Baevski, Wei-Ning Hsu, Qiantong Xu, Arun Babu, Jiatao Gu, and Michael Auli.
\newblock Data2vec: A general framework for self-supervised learning in speech, vision and language.
\newblock \emph{arXiv preprint arXiv:2202.03555}, 2022{\natexlab{b}}.

\bibitem[Bao et~al.(2021)Bao, Dong, and Wei]{bao2021beit}
Hangbo Bao, Li~Dong, and Furu Wei.
\newblock Beit: Bert pre-training of image transformers.
\newblock \emph{arXiv preprint arXiv:2106.08254}, 2021.

\bibitem[Berkes and Wiskott(2005)]{berkes2005slow}
Pietro Berkes and Laurenz Wiskott.
\newblock Slow feature analysis yields a rich repertoire of complex cell properties.
\newblock \emph{Journal of vision}, 5\penalty0 (6):\penalty0 9--9, 2005.

\bibitem[Caron et~al.(2020)Caron, Misra, Mairal, Goyal, Bojanowski, and Joulin]{caron2020unsupervised}
Mathilde Caron, Ishan Misra, Julien Mairal, Priya Goyal, Piotr Bojanowski, and Armand Joulin.
\newblock Unsupervised learning of visual features by contrasting cluster assignments.
\newblock \emph{arXiv preprint arXiv:2006.09882}, 2020.

\bibitem[Caron et~al.(2021)Caron, Touvron, Misra, J{\'e}gou, Mairal, Bojanowski, and Joulin]{caron2021emerging}
Mathilde Caron, Hugo Touvron, Ishan Misra, Herv{\'e} J{\'e}gou, Julien Mairal, Piotr Bojanowski, and Armand Joulin.
\newblock Emerging properties in self-supervised vision transformers.
\newblock \emph{arXiv preprint arXiv:2104.14294}, 2021.

\bibitem[Chen et~al.(2020)Chen, Kornblith, Norouzi, and Hinton]{chen2020simple}
Ting Chen, Simon Kornblith, Mohammad Norouzi, and Geoffrey Hinton.
\newblock A simple framework for contrastive learning of visual representations.
\newblock \emph{preprint arXiv:2002.05709}, 2020.

\bibitem[Chen et~al.(2022)Chen, Ding, Wang, Xin, Mo, Wang, Han, Luo, Zeng, and Wang]{chen2022context}
Xiaokang Chen, Mingyu Ding, Xiaodi Wang, Ying Xin, Shentong Mo, Yunhao Wang, Shumin Han, Ping Luo, Gang Zeng, and Jingdong Wang.
\newblock Context autoencoder for self-supervised representation learning.
\newblock \emph{arXiv preprint arXiv:2202.03026}, 2022.

\bibitem[Chen et~al.(2021)Chen, Xie, and He]{chen2021empirical}
Xinlei Chen, Saining Xie, and Kaiming He.
\newblock An empirical study of training self-supervised vision transformers.
\newblock \emph{arXiv preprint arXiv:2104.02057}, 2021.

\bibitem[Cherti et~al.(2023)Cherti, Beaumont, Wightman, Wortsman, Ilharco, Gordon, Schuhmann, Schmidt, and Jitsev]{cherti2023reproducible}
Mehdi Cherti, Romain Beaumont, Ross Wightman, Mitchell Wortsman, Gabriel Ilharco, Cade Gordon, Christoph Schuhmann, Ludwig Schmidt, and Jenia Jitsev.
\newblock Reproducible scaling laws for contrastive language-image learning.
\newblock In \emph{Proceedings of the IEEE/CVF Conference on Computer Vision and Pattern Recognition}, pages 2818--2829, 2023.

\bibitem[Dogus~Cubuk et~al.(2019)Dogus~Cubuk, Zoph, Mane, and V.~Le]{cubuk2019auto}
Ekin Dogus~Cubuk, Barret Zoph, Vijay Mane, Dandelion~andVasudevan, and Quoc V.~Le.
\newblock Autoaugment: Learning augmentation policies from data.
\newblock In \emph{Proceedings of the IEEE Conference on Computer Vision and Pattern Recognition}, 2019.

\bibitem[Dosovitskiy et~al.(2020)Dosovitskiy, Beyer, Kolesnikov, Weissenborn, Zhai, Unterthiner, Dehghani, Minderer, Heigold, Gelly, et~al.]{dosovitskiy2020image}
Alexey Dosovitskiy, Lucas Beyer, Alexander Kolesnikov, Dirk Weissenborn, Xiaohua Zhai, Thomas Unterthiner, Mostafa Dehghani, Matthias Minderer, Georg Heigold, Sylvain Gelly, et~al.
\newblock An image is worth 16x16 words: Transformers for image recognition at scale.
\newblock \emph{arXiv preprint arXiv:2010.11929}, 2020.

\bibitem[Feichtenhofer et~al.(2021)Feichtenhofer, Fan, Xiong, Girshick, and He]{feichtenhofer2021large}
Christoph Feichtenhofer, Haoqi Fan, Bo~Xiong, Ross Girshick, and Kaiming He.
\newblock A large-scale study on unsupervised spatiotemporal representation learning.
\newblock \emph{Proceedings of the IEEE conference on computer vision and pattern recognition}, 2021.

\bibitem[Feichtenhofer et~al.(2022)Feichtenhofer, Li, He, et~al.]{feichtenhofer2022masked}
Christoph Feichtenhofer, Yanghao Li, Kaiming He, et~al.
\newblock Masked autoencoders as spatiotemporal learners.
\newblock \emph{Advances in neural information processing systems}, 35:\penalty0 35946--35958, 2022.

\bibitem[Field(1994)]{field1994goal}
David~J Field.
\newblock What is the goal of sensory coding?
\newblock \emph{Neural computation}, 6\penalty0 (4):\penalty0 559--601, 1994.

\bibitem[Gidaris et~al.(2020)Gidaris, Bursuc, Komodakis, P{\'e}rez, and Cord]{gidaris2020learning}
Spyros Gidaris, Andrei Bursuc, Nikos Komodakis, Patrick P{\'e}rez, and Matthieu Cord.
\newblock Learning representations by predicting bags of visual words.
\newblock In \emph{Proceedings of the IEEE/CVF Conference on Computer Vision and Pattern Recognition}, pages 6928--6938, 2020.

\bibitem[Girdhar and Grauman(2021)]{girdhar2021anticipative}
Rohit Girdhar and Kristen Grauman.
\newblock Anticipative video transformer.
\newblock In \emph{Proceedings of the IEEE/CVF international conference on computer vision}, pages 13505--13515, 2021.

\bibitem[Girdhar et~al.(2023)Girdhar, El-Nouby, Singh, Alwala, Joulin, and Misra]{girdhar2023omnimae}
Rohit Girdhar, Alaaeldin El-Nouby, Mannat Singh, Kalyan~Vasudev Alwala, Armand Joulin, and Ishan Misra.
\newblock Omnimae: Single model masked pretraining on images and videos.
\newblock In \emph{Proceedings of the IEEE/CVF Conference on Computer Vision and Pattern Recognition}, pages 10406--10417, 2023.

\bibitem[Goroshin et~al.(2015)Goroshin, Bruna, Tompson, Eigen, and LeCun]{goroshin2015unsupervised}
Ross Goroshin, Joan Bruna, Jonathan Tompson, David Eigen, and Yann LeCun.
\newblock Unsupervised learning of spatiotemporally coherent metrics.
\newblock In \emph{Proceedings of the IEEE international conference on computer vision}, pages 4086--4093, 2015.

\bibitem[Goyal et~al.(2017)Goyal, Ebrahimi~Kahou, Michalski, Materzynska, Westphal, Kim, Haenel, Fruend, Yianilos, Mueller-Freitag, et~al.]{goyal2017something}
Raghav Goyal, Samira Ebrahimi~Kahou, Vincent Michalski, Joanna Materzynska, Susanne Westphal, Heuna Kim, Valentin Haenel, Ingo Fruend, Peter Yianilos, Moritz Mueller-Freitag, et~al.
\newblock The" something something" video database for learning and evaluating visual common sense.
\newblock In \emph{Proceedings of the IEEE international conference on computer vision}, pages 5842--5850, 2017.

\bibitem[Grill et~al.(2020)Grill, Strub, Altch{\'e}, Tallec, Richemond, Buchatskaya, Doersch, Pires, Guo, Azar, et~al.]{grill2020bootstrap}
Jean-Bastien Grill, Florian Strub, Florent Altch{\'e}, Corentin Tallec, Pierre~H Richemond, Elena Buchatskaya, Carl Doersch, Bernardo~Avila Pires, Zhaohan~Daniel Guo, Mohammad~Gheshlaghi Azar, et~al.
\newblock Bootstrap your own latent: A new approach to self-supervised learning.
\newblock \emph{arXiv preprint arXiv:2006.07733}, 2020.

\bibitem[Gu et~al.(2018)Gu, Sun, Ross, Vondrick, Pantofaru, Li, Vijayanarasimhan, Toderici, Ricco, Sukthankar, et~al.]{gu2018ava}
Chunhui Gu, Chen Sun, David~A Ross, Carl Vondrick, Caroline Pantofaru, Yeqing Li, Sudheendra Vijayanarasimhan, George Toderici, Susanna Ricco, Rahul Sukthankar, et~al.
\newblock Ava: A video dataset of spatio-temporally localized atomic visual actions.
\newblock In \emph{Proceedings of the IEEE conference on computer vision and pattern recognition}, pages 6047--6056, 2018.

\bibitem[Gupta et~al.(2023)Gupta, Wu, Deng, and Fei-Fei]{gupta2023siamese}
Agrim Gupta, Jiajun Wu, Jia Deng, and Li~Fei-Fei.
\newblock Siamese masked autoencoders.
\newblock \emph{arXiv preprint arXiv:2305.14344}, 2023.

\bibitem[Gutmann and Hyv{\"a}rinen(2012)]{gutmann2012noise}
Michael~U Gutmann and Aapo Hyv{\"a}rinen.
\newblock Noise-contrastive estimation of unnormalized statistical models, with applications to natural image statistics.
\newblock \emph{Journal of machine learning research}, 13\penalty0 (2), 2012.

\bibitem[Han et~al.(2019)Han, Xie, and Zisserman]{han2019video}
Tengda Han, Weidi Xie, and Andrew Zisserman.
\newblock Video representation learning by dense predictive coding.
\newblock In \emph{Proceedings of the IEEE/CVF International Conference on Computer Vision Workshops}, pages 0--0, 2019.

\bibitem[Han et~al.(2020)Han, Xie, and Zisserman]{han2020memory}
Tengda Han, Weidi Xie, and Andrew Zisserman.
\newblock Memory-augmented dense predictive coding for video representation learning.
\newblock In \emph{European conference on computer vision}, pages 312--329. Springer, 2020.

\bibitem[He et~al.(2021)He, Chen, Xie, Li, Doll{\'a}r, and Girshick]{he2021masked}
Kaiming He, Xinlei Chen, Saining Xie, Yanghao Li, Piotr Doll{\'a}r, and Ross Girshick.
\newblock Masked autoencoders are scalable vision learners.
\newblock \emph{arXiv preprint arXiv:2111.06377}, 2021.

\bibitem[Hinton(1989)]{hinton1989connectionist}
Geoffrey~E Hinton.
\newblock Connectionist learning procedures.
\newblock In \emph{Machine learning}, pages 555--610. Elsevier, 1989.

\bibitem[Kalluri et~al.(2023)Kalluri, Pathak, Chandraker, and Tran]{kalluri2023flavr}
Tarun Kalluri, Deepak Pathak, Manmohan Chandraker, and Du~Tran.
\newblock Flavr: Flow-agnostic video representations for fast frame interpolation.
\newblock In \emph{Proceedings of the IEEE/CVF Winter Conference on Applications of Computer Vision}, pages 2071--2082, 2023.

\bibitem[Kaplan et~al.(2020)Kaplan, McCandlish, Henighan, Brown, Chess, Child, Gray, Radford, Wu, and Amodei]{kaplan2020scaling}
Jared Kaplan, Sam McCandlish, Tom Henighan, Tom~B Brown, Benjamin Chess, Rewon Child, Scott Gray, Alec Radford, Jeffrey Wu, and Dario Amodei.
\newblock Scaling laws for neural language models.
\newblock \emph{arXiv preprint arXiv:2001.08361}, 2020.

\bibitem[Kay et~al.(2017)Kay, Carreira, Simonyan, Zhang, Hillier, Vijayanarasimhan, Viola, Green, Back, Natsev, et~al.]{kay2017kinetics}
Will Kay, Joao Carreira, Karen Simonyan, Brian Zhang, Chloe Hillier, Sudheendra Vijayanarasimhan, Fabio Viola, Tim Green, Trevor Back, Paul Natsev, et~al.
\newblock The kinetics human action video dataset.
\newblock \emph{arXiv preprint arXiv:1705.06950}, 2017.

\bibitem[Kayser et~al.(2001)Kayser, Einh{\"a}user, D{\"u}mmer, K{\"o}nig, and K{\"o}rding]{kayser2001extracting}
Christoph Kayser, Wolfgang Einh{\"a}user, Olaf D{\"u}mmer, Peter K{\"o}nig, and Konrad K{\"o}rding.
\newblock Extracting slow subspaces from natural videos leads to complex cells.
\newblock In \emph{Artificial Neural Networks—ICANN 2001: International Conference Vienna, Austria, August 21--25, 2001 Proceedings 11}, pages 1075--1080. Springer, 2001.

\bibitem[Larsson et~al.(2016)Larsson, Maire, and Shakhnarovich]{larsson2016learning}
Gustav Larsson, Michael Maire, and Gregory Shakhnarovich.
\newblock Learning representations for automatic colorization.
\newblock 2016.

\bibitem[Larsson et~al.(2017)Larsson, Maire, and Shakhnarovich]{larsson2017colorization}
Gustav Larsson, Michael Maire, and Gregory Shakhnarovich.
\newblock Colorization as a proxy task for visual understanding.
\newblock 2017.

\bibitem[LeCun(2022)]{lecun2022path}
Yann LeCun.
\newblock A path towards autonomous machine intelligence version 0.9. 2, 2022-06-27.
\newblock 2022.

\bibitem[Lee et~al.(2017)Lee, Huang, Singh, and Yang]{lee2017unsupervised}
Hsin-Ying Lee, Jia-Bin Huang, Maneesh Singh, and Ming-Hsuan Yang.
\newblock Unsupervised representation learning by sorting sequences.
\newblock In \emph{Proceedings of the IEEE international conference on computer vision}, pages 667--676, 2017.

\bibitem[Li et~al.(2022)Li, Wang, Gao, Song, Liu, Li, and Qiao]{li2022uniformer}
Kunchang Li, Yali Wang, Peng Gao, Guanglu Song, Yu~Liu, Hongsheng Li, and Yu~Qiao.
\newblock Uniformer: Unified transformer for efficient spatiotemporal representation learning.
\newblock \emph{arXiv preprint arXiv:2201.04676}, 2022.

\bibitem[Loshchilov and Hutter(2017)]{loshchilov2017decoupled}
Ilya Loshchilov and Frank Hutter.
\newblock Decoupled weight decay regularization.
\newblock \emph{arXiv preprint arXiv:1711.05101}, 2017.

\bibitem[Miech et~al.(2019)Miech, Zhukov, Alayrac, Tapaswi, Laptev, and Sivic]{miech2019howto100m}
Antoine Miech, Dimitri Zhukov, Jean-Baptiste Alayrac, Makarand Tapaswi, Ivan Laptev, and Josef Sivic.
\newblock Howto100m: Learning a text-video embedding by watching hundred million narrated video clips.
\newblock In \emph{Proceedings of the IEEE/CVF international conference on computer vision}, pages 2630--2640, 2019.

\bibitem[Noroozi and Favaro(2016)]{noroozi2016unsupervised}
Mehdi Noroozi and Paolo Favaro.
\newblock Unsupervised learning of visual representations by solving jigsaw puzzles.
\newblock In \emph{European conference on computer vision}, pages 69--84. Springer, 2016.

\bibitem[Oord et~al.(2018)Oord, Li, and Vinyals]{oord2018representation}
Aaron van~den Oord, Yazhe Li, and Oriol Vinyals.
\newblock Representation learning with contrastive predictive coding.
\newblock \emph{arXiv preprint arXiv:1807.03748}, 2018.

\bibitem[Oquab et~al.(2023)Oquab, Darcet, Moutakanni, Vo, Szafraniec, Khalidov, Fernandez, Haziza, Massa, El-Nouby, et~al.]{oquab2023dinov2}
Maxime Oquab, Timoth{\'e}e Darcet, Th{\'e}o Moutakanni, Huy Vo, Marc Szafraniec, Vasil Khalidov, Pierre Fernandez, Daniel Haziza, Francisco Massa, Alaaeldin El-Nouby, et~al.
\newblock Dinov2: Learning robust visual features without supervision.
\newblock \emph{arXiv preprint arXiv:2304.07193}, 2023.

\bibitem[Parthasarathy et~al.(2022)Parthasarathy, Eslami, Carreira, and H{\'e}naff]{parthasarathy2022self}
Nikhil Parthasarathy, SM~Eslami, Jo{\~a}o Carreira, and Olivier~J H{\'e}naff.
\newblock Self-supervised video pretraining yields strong image representations.
\newblock \emph{arXiv preprint arXiv:2210.06433}, 2022.

\bibitem[Pathak et~al.(2016)Pathak, Krahenbuhl, Donahue, Darrell, and Efros]{pathak2016context}
Deepak Pathak, Philipp Krahenbuhl, Jeff Donahue, Trevor Darrell, and Alexei~A Efros.
\newblock Context encoders: Feature learning by inpainting.
\newblock In \emph{Proceedings of the IEEE conference on computer vision and pattern recognition}, pages 2536--2544, 2016.

\bibitem[Pintea et~al.(2014)Pintea, van Gemert, and Smeulders]{pintea2014deja}
Silvia~L Pintea, Jan~C van Gemert, and Arnold~WM Smeulders.
\newblock D{\'e}ja vu: Motion prediction in static images.
\newblock In \emph{Computer Vision--ECCV 2014: 13th European Conference, Zurich, Switzerland, September 6-12, 2014, Proceedings, Part III 13}, pages 172--187. Springer, 2014.

\bibitem[Radford et~al.(2021)Radford, Kim, Hallacy, Ramesh, Goh, Agarwal, Sastry, Askell, Mishkin, Clark, et~al.]{radford2021learning}
Alec Radford, Jong~Wook Kim, Chris Hallacy, Aditya Ramesh, Gabriel Goh, Sandhini Agarwal, Girish Sastry, Amanda Askell, Pamela Mishkin, Jack Clark, et~al.
\newblock Learning transferable visual models from natural language supervision.
\newblock In \emph{International conference on machine learning}, pages 8748--8763. PMLR, 2021.

\bibitem[Rao and Ballard(1999)]{rao1999predictive}
Rajesh~PN Rao and Dana~H Ballard.
\newblock Predictive coding in the visual cortex: a functional interpretation of some extra-classical receptive-field effects.
\newblock \emph{Nature neuroscience}, 2\penalty0 (1):\penalty0 79--87, 1999.

\bibitem[Russakovsky et~al.(2015)Russakovsky, Deng, Su, Krause, Satheesh, Ma, Huang, Karpathy, Khosla, Bernstein, Berg, and Fei-Fei]{russakovsky2015imagenet}
Olga Russakovsky, Jia Deng, Hao Su, Jonathan Krause, Sanjeev Satheesh, Sean Ma, Zhiheng Huang, Andrej Karpathy, Aditya Khosla, Michael Bernstein, Alexander~C. Berg, and Li~Fei-Fei.
\newblock Imagenet large scale visual recognition challenge.
\newblock \emph{International Journal of Computer Vision}, 115\penalty0 (3):\penalty0 211--252, 2015.

\bibitem[Ryali et~al.(2023)Ryali, Hu, Bolya, Wei, Fan, Huang, Aggarwal, Chowdhury, Poursaeed, Hoffman, et~al.]{ryali2023hiera}
Chaitanya Ryali, Yuan-Ting Hu, Daniel Bolya, Chen Wei, Haoqi Fan, Po-Yao Huang, Vaibhav Aggarwal, Arkabandhu Chowdhury, Omid Poursaeed, Judy Hoffman, et~al.
\newblock Hiera: A hierarchical vision transformer without the bells-and-whistles.
\newblock \emph{arXiv preprint arXiv:2306.00989}, 2023.

\bibitem[Sevilla-Lara et~al.(2021)Sevilla-Lara, Zha, Yan, Goswami, Feiszli, and Torresani]{sevilla2021only}
Laura Sevilla-Lara, Shengxin Zha, Zhicheng Yan, Vedanuj Goswami, Matt Feiszli, and Lorenzo Torresani.
\newblock Only time can tell: Discovering temporal data for temporal modeling.
\newblock In \emph{Proceedings of the IEEE/CVF winter conference on applications of computer vision}, pages 535--544, 2021.

\bibitem[Spelke et~al.(1995)Spelke, Vishton, and Von~Hofsten]{spelke1995object}
Elizabeth~S Spelke, Peter Vishton, and Claes Von~Hofsten.
\newblock Object perception, object-directed action, and physical knowledge in infancy.
\newblock 1995.

\bibitem[Srivastava et~al.(2015)Srivastava, Mansimov, and Salakhudinov]{srivastava2015unsupervised}
Nitish Srivastava, Elman Mansimov, and Ruslan Salakhudinov.
\newblock Unsupervised learning of video representations using lstms.
\newblock In \emph{International conference on machine learning}, pages 843--852. PMLR, 2015.

\bibitem[Sun et~al.(2019)Sun, Myers, Vondrick, Murphy, and Schmid]{sun2019videobert}
Chen Sun, Austin Myers, Carl Vondrick, Kevin Murphy, and Cordelia Schmid.
\newblock Videobert: A joint model for video and language representation learning.
\newblock In \emph{Proceedings of the IEEE/CVF international conference on computer vision}, pages 7464--7473, 2019.

\bibitem[Sur{\'\i}s et~al.(2021)Sur{\'\i}s, Liu, and Vondrick]{suris2021learning}
D{\'\i}dac Sur{\'\i}s, Ruoshi Liu, and Carl Vondrick.
\newblock Learning the predictability of the future.
\newblock In \emph{Proceedings of the IEEE/CVF Conference on Computer Vision and Pattern Recognition}, pages 12607--12617, 2021.

\bibitem[Tan et~al.(2023)Tan, De~Lange, Iuzzolino, Plummer, Saenko, Ridgeway, and Torresani]{tan2023multiscale}
Reuben Tan, Matthias De~Lange, Michael Iuzzolino, Bryan~A Plummer, Kate Saenko, Karl Ridgeway, and Lorenzo Torresani.
\newblock Multiscale video pretraining for long-term activity forecasting.
\newblock \emph{arXiv preprint arXiv:2307.12854}, 2023.

\bibitem[Tarvainen and Valpola(2017)]{tarvainen2017mean}
Antti Tarvainen and Harri Valpola.
\newblock Mean teachers are better role models: Weight-averaged consistency targets improve semi-supervised deep learning results.
\newblock \emph{arXiv preprint arXiv:1703.01780}, 2017.

\bibitem[Tian et~al.(2021)Tian, Chen, and Ganguli]{tian2021understanding}
Yuandong Tian, Xinlei Chen, and Surya Ganguli.
\newblock Understanding self-supervised learning dynamics without contrastive pairs.
\newblock In \emph{International Conference on Machine Learning}, pages 10268--10278. PMLR, 2021.

\bibitem[Tong et~al.(2022)Tong, Song, Wang, and Wang]{tong2022videomae}
Zhan Tong, Yibing Song, Jue Wang, and Limin Wang.
\newblock Videomae: Masked autoencoders are data-efficient learners for self-supervised video pre-training.
\newblock \emph{Advances in neural information processing systems}, 35:\penalty0 10078--10093, 2022.

\bibitem[Van~Horn et~al.(2018)Van~Horn, Mac~Aodha, Song, Cui, Sun, Shepard, Adam, Perona, and Belongie]{van2018inaturalist}
Grant Van~Horn, Oisin Mac~Aodha, Yang Song, Yin Cui, Chen Sun, Alex Shepard, Hartwig Adam, Pietro Perona, and Serge Belongie.
\newblock The inaturalist species classification and detection dataset.
\newblock In \emph{Proceedings of the IEEE conference on computer vision and pattern recognition}, pages 8769--8778, 2018.

\bibitem[Vincent et~al.(2008)Vincent, Larochelle, Bengio, and Manzagol]{denoising_vincent}
Pascal Vincent, Hugo Larochelle, Yoshua Bengio, and Pierre-Antoine Manzagol.
\newblock Extracting and composing robust features with denoising autoencoders.
\newblock In \emph{Proceedings of the 25th International Conference on Machine Learning}, ICML '08, page 1096–1103, 2008.

\bibitem[Vincent et~al.(2010)Vincent, Larochelle, Lajoie, Bengio, Manzagol, and Bottou]{vincent2010stacked}
Pascal Vincent, Hugo Larochelle, Isabelle Lajoie, Yoshua Bengio, Pierre-Antoine Manzagol, and L{\'e}on Bottou.
\newblock Stacked denoising autoencoders: Learning useful representations in a deep network with a local denoising criterion.
\newblock \emph{Journal of machine learning research}, 11\penalty0 (12), 2010.

\bibitem[Vondrick et~al.(2016)Vondrick, Pirsiavash, and Torralba]{vondrick2016anticipating}
Carl Vondrick, Hamed Pirsiavash, and Antonio Torralba.
\newblock Anticipating visual representations from unlabeled video.
\newblock In \emph{Proceedings of the IEEE conference on computer vision and pattern recognition}, pages 98--106, 2016.

\bibitem[Wang et~al.(2010)Wang, Li, and Konig]{wang2010learning}
Fei Wang, Ping Li, and Arnd~Christian Konig.
\newblock Learning a bi-stochastic data similarity matrix.
\newblock In \emph{2010 IEEE International Conference on Data Mining}, pages 551--560. IEEE, 2010.

\bibitem[Wang et~al.(2023{\natexlab{a}})Wang, Huang, Zhao, Tong, He, Wang, Wang, and Qiao]{wang2023videomae}
Limin Wang, Bingkun Huang, Zhiyu Zhao, Zhan Tong, Yinan He, Yi~Wang, Yali Wang, and Yu~Qiao.
\newblock Videomae v2: Scaling video masked autoencoders with dual masking.
\newblock In \emph{Proceedings of the IEEE/CVF Conference on Computer Vision and Pattern Recognition}, pages 14549--14560, 2023{\natexlab{a}}.

\bibitem[Wang et~al.(2023{\natexlab{b}})Wang, Chen, Wu, Chen, Dai, Liu, Yuan, and Jiang]{wang2023masked}
Rui Wang, Dongdong Chen, Zuxuan Wu, Yinpeng Chen, Xiyang Dai, Mengchen Liu, Lu~Yuan, and Yu-Gang Jiang.
\newblock Masked video distillation: Rethinking masked feature modeling for self-supervised video representation learning.
\newblock In \emph{Proceedings of the IEEE/CVF Conference on Computer Vision and Pattern Recognition}, pages 6312--6322, 2023{\natexlab{b}}.

\bibitem[Wang et~al.(2022)Wang, Li, Li, He, Huang, Zhao, Zhang, Xu, Liu, Wang, et~al.]{wang2022internvideo}
Yi~Wang, Kunchang Li, Yizhuo Li, Yinan He, Bingkun Huang, Zhiyu Zhao, Hongjie Zhang, Jilan Xu, Yi~Liu, Zun Wang, et~al.
\newblock Internvideo: General video foundation models via generative and discriminative learning.
\newblock \emph{arXiv preprint arXiv:2212.03191}, 2022.

\bibitem[Wiskott and Sejnowski(2002)]{wiskott2002slow}
Laurenz Wiskott and Terrence~J Sejnowski.
\newblock Slow feature analysis: Unsupervised learning of invariances.
\newblock \emph{Neural computation}, 14\penalty0 (4):\penalty0 715--770, 2002.

\bibitem[Wu et~al.(2018)Wu, Xiong, Yu, and Lin]{wu2018unsupervised}
Zhirong Wu, Yuanjun Xiong, Stella~X Yu, and Dahua Lin.
\newblock Unsupervised feature learning via non-parametric instance discrimination.
\newblock In \emph{Proceedings of the IEEE conference on computer vision and pattern recognition}, pages 3733--3742, 2018.

\bibitem[Xie et~al.(2021)Xie, Zhang, Cao, Lin, Bao, Yao, Dai, and Hu]{xie2021simmim}
Zhenda Xie, Zheng Zhang, Yue Cao, Yutong Lin, Jianmin Bao, Zhuliang Yao, Qi~Dai, and Han Hu.
\newblock Simmim: A simple framework for masked image modeling.
\newblock \emph{arXiv preprint arXiv:2111.09886}, 2021.

\bibitem[Xu et~al.(2019)Xu, Xiao, Zhao, Shao, Xie, and Zhuang]{xu2019self}
Dejing Xu, Jun Xiao, Zhou Zhao, Jian Shao, Di~Xie, and Yueting Zhuang.
\newblock Self-supervised spatiotemporal learning via video clip order prediction.
\newblock In \emph{Proceedings of the IEEE/CVF Conference on Computer Vision and Pattern Recognition}, pages 10334--10343, 2019.

\bibitem[Xu et~al.(2021)Xu, Ghosh, Huang, Okhonko, Aghajanyan, Metze, Zettlemoyer, and Feichtenhofer]{xu2021videoclip}
Hu~Xu, Gargi Ghosh, Po-Yao Huang, Dmytro Okhonko, Armen Aghajanyan, Florian Metze, Luke Zettlemoyer, and Christoph Feichtenhofer.
\newblock Videoclip: Contrastive pre-training for zero-shot video-text understanding.
\newblock \emph{arXiv preprint arXiv:2109.14084}, 2021.

\bibitem[Yu et~al.(2022)Yu, Wang, Vasudevan, Yeung, Seyedhosseini, and Wu]{yu2022coca}
Jiahui Yu, Zirui Wang, Vijay Vasudevan, Legg Yeung, Mojtaba Seyedhosseini, and Yonghui Wu.
\newblock Coca: Contrastive captioners are image-text foundation models.
\newblock \emph{arXiv preprint arXiv:2205.01917}, 2022.

\bibitem[Yuan et~al.(2023)Yuan, Gundavarapu, Zhao, Zhou, Cui, Jiang, Yang, Jia, Weyand, Friedman, et~al.]{yuan2023videoglue}
Liangzhe Yuan, Nitesh~Bharadwaj Gundavarapu, Long Zhao, Hao Zhou, Yin Cui, Lu~Jiang, Xuan Yang, Menglin Jia, Tobias Weyand, Luke Friedman, et~al.
\newblock Videoglue: Video general understanding evaluation of foundation models.
\newblock \emph{arXiv preprint arXiv:2307.03166}, 2023.

\bibitem[Zellers et~al.(2022)Zellers, Lu, Lu, Yu, Zhao, Salehi, Kusupati, Hessel, Farhadi, and Choi]{zellers2022merlot}
Rowan Zellers, Jiasen Lu, Ximing Lu, Youngjae Yu, Yanpeng Zhao, Mohammadreza Salehi, Aditya Kusupati, Jack Hessel, Ali Farhadi, and Yejin Choi.
\newblock Merlot reserve: Neural script knowledge through vision and language and sound.
\newblock In \emph{Proceedings of the IEEE/CVF Conference on Computer Vision and Pattern Recognition}, pages 16375--16387, 2022.

\bibitem[Zhou et~al.(2014)Zhou, Lapedriza, Xiao, Torralba, and Oliva]{places205}
Bolei Zhou, Agata Lapedriza, Jianxiong Xiao, Antonio Torralba, and Aude Oliva.
\newblock Learning deep features for scene recognition using places database.
\newblock In Z.~Ghahramani, M.~Welling, C.~Cortes, N.~Lawrence, and K.Q. Weinberger, editors, \emph{Advances in Neural Information Processing Systems}, volume~27. Curran Associates, Inc., 2014.
\newblock \url{https://proceedings.neurips.cc/paper/2014/file/3fe94a002317b5f9259f82690aeea4cd-Paper.pdf}.

\bibitem[Zhou et~al.(2021)Zhou, Wei, Wang, Shen, Xie, Yuille, and Kong]{zhou2021ibotyes}
Jinghao Zhou, Chen Wei, Huiyu Wang, Wei Shen, Cihang Xie, Alan Yuille, and Tao Kong.
\newblock Ibot: Image bert pre-training with online tokenizer.
\newblock \emph{arXiv preprint arXiv:2111.07832}, 2021.

\bibitem[Zou et~al.(2012)Zou, Zhu, Yu, and Ng]{zou2012deep}
Will Zou, Shenghuo Zhu, Kai Yu, and Andrew Ng.
\newblock Deep learning of invariant features via simulated fixations in video.
\newblock \emph{Advances in neural information processing systems}, 25, 2012.

\end{thebibliography}

\clearpage
\newpage

\onecolumn

\beginappendix

\section{Extended Related Works}

We first review approaches for learning visual perception from static images before discussing strategies for learning from video.

\subsection*{Weakly-Supervised Learning from Static Images}
One family of approaches for learning visual perception from static images trains a visual encoder to predict the representations of text captions often found accompanying images from the Web, as in CLIP~\citep{radford2021learning} or CoCa~\citep{yu2022coca}.
The largest open source CLIP model to date, numbering 2B parameters and trained on over 2B web-scraped images~\citep{cherti2023reproducible}, demonstrates impressive performance on a wide range of downstream image and video tasks.
Notably, this is achieved using only the light-weight adaptation of task-specific heads, also referred to as frozen-evaluation, and does not require expensive end-to-end fine-tuning of the pretrained model.

\subsection*{Self-Supervised Learning from Static Images}
Other approaches for learning from static images leverage unsupervised objectives.
Initial works on self-supervised approaches are based on sparse coding or hand-crafted pretext tasks, such as colorization~\citep{larsson2016learning,larsson2017colorization}, rotation prediction~\citep{gidaris2020learning}, and jigsaws~\citep{noroozi2016unsupervised}.
More recent approaches leverage invariance-based objectives by training a visual encoder to be invariant to hand-crafted image transformations~\citep{wu2018unsupervised,chen2020simple}.

Another family of methods learn representations using denoising autoencoders~\citep{denoising_vincent}; image inpainting is one popular instantiation of this idea~\citep{pathak2016context}.
More recently, masked autoencoders~\citep{he2021masked} train an encoder-decoder transformer to predict missing pixels of a masked image.
Follow-up work addresses the indeterminism of pixel reconstruction by exploring instantiations of masked image modeling in latent space~\citep{baevski2022data2vec,assran2023self,baevski2022efficient}.
These approaches can be seen as applications of the predictive feature principle in the image modality.

There are also various methods that combine both masked image modeling and invariance criteria to learn visual representations from static images, such as iBOT~\citep{zhou2021ibotyes} and DINOv2~\citep{zhou2021ibotyes, oquab2023dinov2}, the latter is currently the most competitive instantiation of self-supervised learning with static images, scaled to a model with over 1.1B parameters trained on a curated dataset of 142M images.

\subsection*{Weakly-Supervised Learning from Videos}
One family of approaches for learning visual perception from videos relies on weakly-supervised guidance from closed captioning, often computed from an ASR transcription of audio data accompanying internet videos.
For instance, VideoBERT~\citep{sun2019videobert,xu2021videoclip} trains a video encoder to predict masked spans in the textual closed captions.
Similarly, VideoCLIP~\citep{xu2021videoclip} trains a video encoder to predict the representation of video captions computed by a text encoder.
Follow-up work such as MERLOT~\citep{zellers2022merlot}, VATT~\citep{akbari2021vatt}, and InternVideo~\citep{wang2022internvideo} extended VideoCLIP by incorporating additional unsupervised objectives.

\subsection*{Self-Supervised Learning from Videos}
Similar to unsupervised learning from images, a family of unsupervised video representation learning approaches enforces a spatio-temporal representation of a video clip to be invariant to hand-crafted spatio-temporal data augmentations~\citep{parthasarathy2022self}.
However, one obvious insight is that the temporal ordering of visual information in video can provide implicit supervision.
Indeed, this insight is the key insight leveraged by many works on unsupervised video learning.
Towards leveraging temporal information as supervision, some approaches train a visual encoder by predicting the temporal ordering of frames~\citep{xu2019self, lee2017unsupervised}.
Other approaches seek to predict low-level motion vectors computed from optical flow~\citep{pintea2014deja}, or to predict mixing pixels in video frames, using either a frame-interpolation objective~\citep{kalluri2023flavr} or a denoising autoencoder~\citep{tong2022videomae, feichtenhofer2022masked, wang2023videomae}.

\section{Extended Description of V-JEPA}
\label{appendix:vjepa_extended_description}

In this section, we provide an in-depth description of our approach \putalg that is illustrated in Figure~\ref{fig:vjepa-complex}.

\begin{figure}[h]
    \centering
    \includegraphics[width=0.9\linewidth]{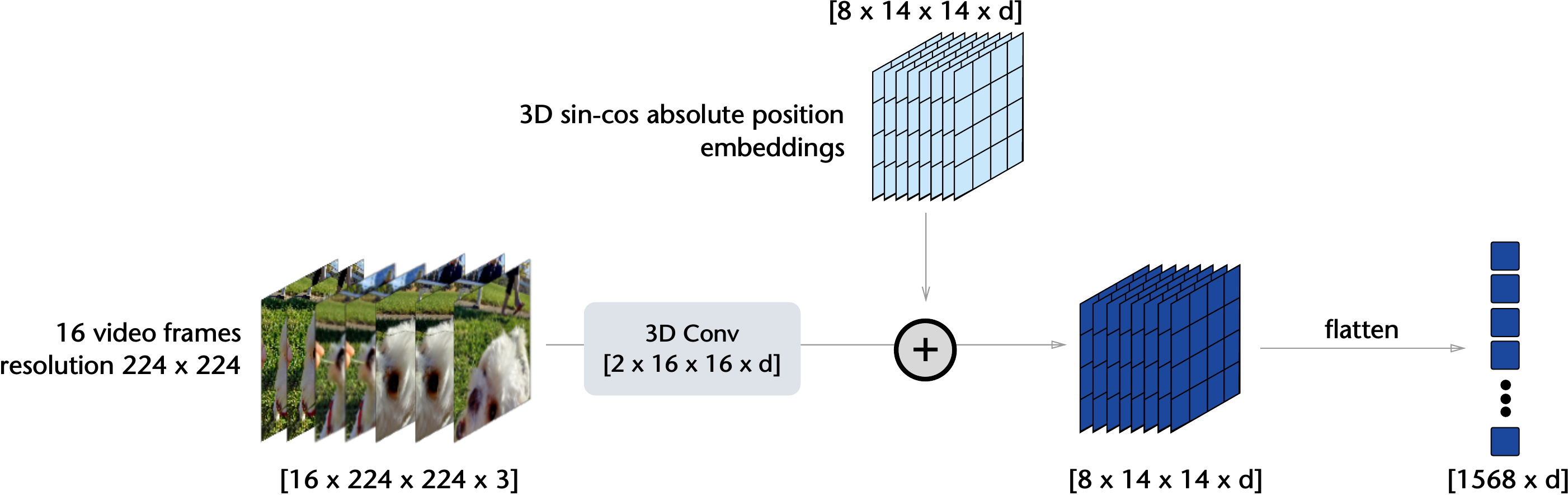}
    \caption{\small{\bf \putalg} training operates on a video clip flattened into a sequence of tokens. To convert a video clip of size $16 \times 224 \times 224 \times 3$ into a 1D token sequence, we apply a 3D convolution comprising $d$ filters of size $2 \times 16 \times 16$ with a temporal stride of $2$ and a spatial stride of $16$, resulting in a tensor of shape $8 \times 14 \times 14 \times d$.
Next we add absolute 3D sin-cos positional embeddings to the spatio-temporal feature map and flatten it, resulting in a 1D token sequence of shape $1568 \times d$.}
    \label{fig:patchitfy}
\end{figure}

\paragraph{\bf Input.}
Unless stated otherwise, during during pretraining, we always randomly sample a clip of 16 frames from each input video with a temporal stride of 4 between sampled frames.
An input video clip therefore covers 64 frames in total, or roughly 2 seconds of a given video running at 30 frames per second.
We then resize the video's spatial dimensions to $224 \times 224$, resulting in an overall shape of $16 \times 224 \times 224 \times 3$ for the entire clip.
Since ViT networks process a 1D sequence of tokens, we must convert an input video clip into a 1D token sequence.
To do so, we apply a 3D convolution comprising $d$ filters of size $2 \times 16 \times 16$ with a temporal stride of $2$ and a spatial stride of $16$, resulting in a tensor of shape $8 \times 14 \times 14 \times d$.
Next we add absolute 3D sin-cos positional embeddings to the spatio-temporal feature map and flatten it, resulting in a 1D token sequence of shape $1568 \times d$.
This process is demonstrated in Figure~\ref{fig:patchitfy}.

\paragraph{\bf \putalg.}
We sample both a video clip, and a video mask in each iteration.
We denote a video clip represented as a 1D token sequence of length $L=1568$ by $x_{L} = (x_1, \ldots, x_L)$.
Similarly, given a mask of $M < L$ patches, leaving $N=L-M$ patches unmasked, we denote the indices of masked patches by $(i_1, \ldots, i_M)$ and its complement (the indices of unmasked patches) by $(j_1, \ldots, j_{N})$.

{\bf \it Computing the  $x$-representations.}
To compute the \putalg loss, we first produce the $x$-representations by masking the video clip and feeding it into the $x$-encoder; we denote the masked video by $x_N = (x_{j_1}, \ldots, x_{j_N})$.
Applying the $x$-encoder $E_\theta(\cdot)$ to the masked clip gives a sequence of patch representations, denoted as
$z_N = E_{\theta}(x_N) = (z_{j_1}, \ldots, z_{j_N}).$

{\bf \it Predicting the target.}
Next, the \putalg predictor network $P_\phi(\cdot, \cdot)$ takes as input the tokens produced by the $x$-encoder and predicts the missing regions in the video clip, which are specified by a set of learnable mask tokens.
Specifically, the mask tokens are parameterized as the sum of a shared learnable vector and an absolute 3D sin-cos positional embedding, denoted by $m_M = (m_{i_1}, \ldots, m_{i_M})$.
The output of the predictor is thus given by,
$\hat s_M = P_\phi(z_N, m_M) = (\hat s_{i_1}, \ldots, \hat s_{i_M}),$
corresponding to a $d$-dimensional output for each of the $M$ masked patches.

{\it Computing the $y$-representations.}
Finally to compute the prediction targets, the entire unmasked video clip is processed by the $y$-encoder to obtain a set of target representations, denoted by
$s_L = \overline{E}_{\theta}(x_L) = (s_1, \ldots, s_L).$
The \putalg loss is now computed as
\begin{equation}    
    \label{eq:loss_detail}
    \text{Loss} = \frac{1}{M} \sum_{k \in (i_1, \ldots, i_M)} \lVert \hat s_{k} - s_k \rVert_1,
\end{equation}
which is simply the average $L_1$ distance between the output of the predictor and the $y$-encoder.
We then compute a gradient update with respect to the parameters of the $x$-encoder, $\theta$, and the predictor, $\phi$, and subsequently update the parameters of the $y$-encoder as an exponential moving average of the context encoder weights (Polyak average).

\paragraph{\bf Multi-Mask Prediction.}
To increase the efficiency of \putalg, we use a multi-masking strategy~\citep{caron2020unsupervised,baevski2022efficient}, which enables us to amortize the cost of the target computation.
As mentioned in Section~\ref{sec:methodology}, for a given video clip, we sample 2 different masks, short-range and long-range.
While we need to forward propagate the $x$-encoder and predictor separately for each mask, we only need to compute the $y$-representation once.

\section{Pretraining details} \label{app:pretraining}

\begin{table}
    \centering
    \caption{\bf pretraining hyper-parameters for \putalg.}
    \label{tab:app_hyper}
    {\fontsize{9pt}{9pt}\selectfont
    \setlength{\tabcolsep}{2pt}
    \begin{tabular}{@{} l c c c @{}}
        \toprule
        Hyper-parameter & ViT-L/16$_{224}$ & ViT-H/16$_{224}$ &
        ViT-H/16$_{384}$ \\
        \midrule \textit{data} \\ datasets & VideoMix2M & VideoMix2M &
        VideoMix2M \\ resolution & 224 & 224 & 384 \\ num\_frames & 16 & 16 &
        16 \\ temporal\_stride & 4 & 4 & 4 \\ horizontal\_flip & true & true &
        true \\ random\_resize\_scale & (0.3, 1.0) & (0.3, 1.0) & (0.3, 1.0)
        \\ random\_resize\_aspect\_ratio & (0.75, 1.35) & (0.75, 1.35) &
        (0.75, 1.35)\\ \midrule \textit{masking} \\ block\_aspect\_ratio &
        (0.75, 1.5) & (0.75, 1.5) & (0.75,
        1.5)\\ shortrange\_mask\_num\_blocks & 8 & 8 &
        8\\ shortrange\_mask\_spatial\_scale & 0.15 & 0.15 &
        0.15\\ longrange\_mask\_num\_blocks & 2 & 2 &
        2\\ longrange\_mask\_spatial\_scale & 0.7 & 0.7 & 0.7\\ \midrule
        \textit{optimization} \\ batch\_size & 3072 & 3072 & 2400
        \\ total\_number\_of\_iterations & 90000 & 90000 & 90000
        \\ warmup\_iterations & 12000 & 12000 & 12000 \\ lr & 6.25e-4 &
        6.25$\times10^{-4}$ & 6.25$\times10^{-4}$ \\ start\_lr & 2$\times10^{-4}$ & 2$\times10^{-4}$ & 2$\times10^{-4}$ \\ final\_lr &
        1$\times10^{-6}$ & 1$\times10^{-6}$ & 1$\times10^{-6}$ \\ start\_momentum & 0.998 & 0.998 & 0.998
        \\ final\_momentum & 1.0 & 1.0 & 1.0 \\ start\_weight\_decay & 0.04 &
        0.04 & 0.04 \\ final\_weight\_decay & 0.4 & 0.4 & 0.4
        \\ scheduler\_scale\_factor & 1.25 & 1.25 & 1.25 \\

        \midrule \textit{architecture} \\ patch\_size & 16 & 16 & 16
        \\ tubelet\_size & 2 & 2 & 2 \\ pred\_depth & 12 & 12 & 12
        \\ pred\_embed\_dim & 384 & 384 & 384 \\
        
        \midrule \textit{hardware} \\ dtype & bfloat16 & bfloat16 & bfloat16
        \\ accelerator & A100 80G & A100 80G & A100 80G \\
        \bottomrule
    \end{tabular}}
\end{table}

In section, we report \putalg pretraining details. Table~\ref{tab:app_hyper} summarizes the main hyperparameters used during pretraining.

\paragraph{Architectures.}
We use Vision Transformer~\citep{dosovitskiy2020image} (ViT) architectures for the $x$-encoder and $y$-encoder.
We train three \putalg encoders: a ViT-L/16$_{224}$, a ViT-H/16$_{224}$ and a ViT-H/16$_{384}$. All three encoders take as input a short video clip of 16 frames with a temporal stride of 4 between consecutive frames. 
The subscripts, $224$ and $384$, indicate the spatial resolution of the video clip.
\putalg flattens the video clip into a sequence of non-overlapping spatio-temporal patches of size $16 \times 16\times 2$ (see Figure~\ref{fig:patchitfy}).
For all three models, the predictor is designed as a narrow ViT architecture, consisting of 12 transformer blocks with an embedding dimension of 384.
For simplicity, we keep the number of self-attention heads in the predictor equal to that of the backbone used for the context-encoder/target-encoder.
\putalg is pretrained \emph{without} using a {\tt [cls]} token.

\paragraph{Optimization.}
We use AdamW~\citep{loshchilov2017decoupled} to optimize the $x$-encoder and predictor weights.
The ViT-L/16$_{224}$ and ViT-H/16$_{224}$ models use a batch size of $3072$ while the ViT-H/16$_{384}$ uses a batch size of $2400$.
Models are trained for a total of 90,000 iterations.
The learning rate is linearly increased from $2\times 10^{-4}$ to $6.25\times 10^{-4}$ during the first $12,000$ iterations of pretraining, and decayed to $10^{-6}$ following a cosine schedule.
Weight-decay is also linearly increased from $0.04$ to $0.4$ throughout pretraining.
The $y$-encoder weights are initialized identically to the $x$-encoder, and subsequently updated as an exponential moving average (EMA)~\citep{tarvainen2017mean} of the $x$-encoder weights using a momentum value which starts at $0.998$ and is linearly increased to $1.0$ during training~\citep{caron2021emerging, assran2022masked}.
We scale all hyper-parameter schedules 25\% beyond the actual training schedule.
Specifically, the learning rate schedule, weight-decay schedule, and EMA schedule are computed assuming a training length of 112,500 iterations, even though we only train our model for 90,000 iterations.
We found the last $25\%$ of the default scheduler period to update hyper-parameters too aggressively, and simply truncating the schedulers improved performance.

\paragraph{Masking.}
As described in Section~\ref{sec:methodology}, we propose a 3D Multi-Block masking strategy. We use two type of masks: short-range masks, where we take the union of $8$ randomly sampled target blocks with a spatial scale of $0.15$, and long-range masks, where we take the union of $2$ randomly sampled target blocks with a spatial scale of $0.7$. In both cases, the aspect ratio for all sampled blocks is randomly chosen in the range $(0.75, 1.5)$.

\section{Evaluation details} \label{app:evaluation}

\subsection{Frozen classification}

\begin{table}
    \centering
    \caption{\bf\small Frozen Evaluation hyper-parameters.}
    \label{tab:frozen_hp}
    {\fontsize{9pt}{9pt}\selectfont
    \setlength{\tabcolsep}{2pt}
    \begin{tabular}{l c c c c c}
        \toprule
        Hyper-parameter & K400 & SSv2 & IN1K & Place205 & iNat21 \\
        \midrule \textit{data} \\
        num\_clips & 8 & 1 & N.A. & N.A. & N.A. \\
        num\_frames & 16 & 16 & N.A. & N.A. & N.A.\\
        temporal\_stride & 4 & 4 & N.A. & N.A. & N.A. \\
        horizontal\_flip & true & true & true & true & true\\
        random\_resize\_scale & (0.08, 1.0) & (0.08, 1.0) & (0.08, 1.0) & (0.08, 1.0) & (0.08, 1.0)\\
        random\_resize\_aspect\_ratio & (0.75, 1.33) & (0.75, 1.33) & (0.75, 1.33) & (0.75, 1.33) & (0.75, 1.33)\\
        auto\_augment & false & false & true & true & true \\
        \midrule \textit{optimization} \\
        batch\_size & 256 & 256 & 1024 & 1024 & 1024 \\
        epochs & 20 & 20 & 20 & 20 & 20\\
        lr & 1e-3 & 1e-3 & 1e-3 & 1e-3 & 1e-3  \\
        final\_lr & 0 & 0 & 0 & 0 & 0\\
        weight\_decay & 0.01 & 0.01 & 0.01 & 0.01 & 0.01 \\\bottomrule
    \end{tabular}}
\end{table}

\paragraph{Attentive Probing.}
Given an input video, $x_L$, the \putalg target encoder $\overline{E}_{{\theta}}(\cdot)$ outputs a sequence of $L$ tokens, $E_{\theta}(x_L) = (s_1, \ldots, s_L)$, where $s_i \in \mathbb{R}^d$.
To pool this sequence of tokens into a single feature vector, we apply a lightweight non-linear cross-attention block which replace the self-attention operation of a transformer block with cross attention.
Specifically, the cross-attention performs the following computation:
\[
    \sum^L_{i=1}{\frac{\exp(q^\top {\bf W_k} s_i)}{\sum_j \exp(q^\top {\bf W_k} s_j)} {\bf W_v} s_i },
\]
where $\bf{W_k}, {\bf W_v} \in R^{d \times d}$ are the key and value matrices, and $q \in R^d$ is a learnable query token.
The output of the cross-attention is then added back to the query token (residual connection), and then fed into two-layer MLP with a single GeLU activation, followed by a LayerNorm, and finally a linear classifier.
The parameters of the cross-attention block are jointly learned with that of the linear classifier for the downstream task, while the encoder parameters are kept frozen.
Note that, in practice, we actually use an attentive probe with 12 heads, each of dimension $12$.
In Appendix~\ref{app:extra_results} we show that baselines benefit from the attentive probing protocol.

\paragraph{Optimization.}
For all the tasks, we use AdamW optimizer with a cosine scheduler (no warmup) that decays the learning rate from $0.001$ to $0$.
We use a fixed weight-decay of $0.01$ and apply simple data augmentations (random resized crops and horizontal flips) during training of the attentive probe, except on image tasks, where we apply AutoAugment~\citep{cubuk2019auto}.
Table~\ref{tab:frozen_hp} reports the hyperparameters for each downstream evaluation.

\paragraph{Extension to multiple clips.}
Unless stated otherwise, our attentive probe takes 8 clips of 16 frames as input on Kinetics, and 2 clips of 16 frames on Something-Somethingv2 to increase the temporal coverage of the video.
Specifically,  we first divide a video in 8 (or 2) equal-length temporal segments, and sample 1 clip at random per segment.
The video encoder $\overline{E}_{{\theta }}$ processes each clip separately and produces a clip-level feature map.
The feature maps for each clip are then concatenated together and fed to the attentive probe.
At test time, we average the prediction of 3 spatial views following standard practice in video classification.

\paragraph{Application of video models to images.}
To evaluate the video models on image tasks, we simply duplicate input images to generate still video clips of 16 frames.
We perform this duplication operation simply for convenience in evaluation of the video models, however we find this step to be unnecessary in general.
Given a video tokenizer implemented as a 3D-conv with a temporal stride of $2$, it is sufficient to simply duplicate the image into a 2 frame video clip.
This would result in the same number of input tokens as that produced by a static image model with a 2D-conv tokenizer.

\paragraph{Application of image models to videos.}
To evaluate image models such as DINOv2 and OpenCLIP on video tasks, we simply process each frame independently with the image encoder to produce a frame-level feature map.
The feature maps for each frame are then concatenated and fed to the attentive probe, just as we do with the clip-level feature maps when evaluating video models.

\subsection{Frozen detection}

We evaluate our model on the AVA~\citep{gu2018ava} spatio-temporal localization of human actions dataset, containing 211k training and 57k validation video segments. We follow the experimental protocol of~\citep{feichtenhofer2021large}, and use precomputed masks from a pretrained Faster-RCNN adapted to videos, which uses a ResNeXt-101-FPN backbone and is pretrained on ImageNet and COCO. We train a linear classifier on top of the \textit{frozen} \putalg features to classify the extracted regions of interest and report mean Average Precision (mAP) on the 60 most common classes.
Hyper-parameters are provided in Table~\ref{tab:ava}.
Our frozen features are obtained by concatenating the last layer of the transformer encoder with three intermediate layers.
We use a batch size of 64 and pretrain for 30 epochs with AdamW using a learning rate of 0.0001 with 2 epochs of warmup and a weight decay of 0.05.

\begin{table}
    \centering
    \caption{\bf\small Frozen Detection hyper-parameters.}
    \label{tab:ava}
    {\fontsize{9pt}{9pt}\selectfont
    \setlength{\tabcolsep}{2pt}
    \begin{tabular}{@{} l c c @{}}
        \toprule
        Hyper-parameter & ViT-L/16 & ViT-H/16 \\\midrule
        out\_layers & [18, 20, 22, 24] & [26, 28, 30, 32] \\
        batch\_size & 64 & 64 \\
        epochs & 30 & 30\\
        opt & AdamW & AdamW \\
        opt\_eps & 0.00000001 & 0.00000001 \\
        momentum & 0.9 & 0.9 \\
        weight\_decay & 0.05 & 0.05 \\
        lr & 0.0001 & 0.0001 \\
        warmup\_lr & 0.000001 & 0.000001 \\
        min\_lr & 0.000001 & 0.000001 \\
        warmup\_epochs & 2 & 2 \\
        warmup\_steps & 1 & 1 \\
        \bottomrule
    \end{tabular}}
\end{table}

\subsection{Finetuning}

Following~\citet{tong2022videomae}, we finetune a linear layer on top of our model, using a layer decay schema and mixup as the data augmentation pipeline. We provide all hyper-parameters for both K400 and SSv2 in Table~\ref{tab:fine_tuning_hp}.

\begin{table}
    \centering
    \caption{\bf\small Finetuning Evaluation hyper-parameters.}
    \label{tab:fine_tuning_hp}
    {\fontsize{10pt}{10pt}\selectfont
    \setlength{\tabcolsep}{2pt}
    \begin{tabular}{ l cc @{\hspace{1cm}} cc}
        \toprule
        Hyper-parameter & \multicolumn{2}{c}{K400} & \multicolumn{2}{c}{SSv2} \\
        \midrule \textit{data} \\
        num\_segments & \multicolumn{4}{c}{1}\\
        num\_frames & \multicolumn{4}{c}{16}  \\
        sampling\_rate & \multicolumn{4}{c}{4} \\
        resolution & \multicolumn{4}{c}{224}  \\
        
        \midrule \textit{model} \\
        model\_name & ViT-L/16 & ViT-H/16 & ViT-L/16 & ViT-H/16 \\
        drop\_path & 0.1 & 0.2 & 0.2 & 0.2 \\
        head\_drop\_rate & 0. & 0. & 0.5 & 0.5 \\
        
        \midrule \textit{optimization} \\
        batch\_size & 256& 1024 & 256 & 256 \\
        epochs & 35 & 25 & 15 & 15 \\
        opt & \multicolumn{4}{c}{adamw} \\
        opt\_eps & \multicolumn{4}{c}{0.00000001}\\
        momentum & \multicolumn{4}{c}{0.9} \\
        weight\_decay & \multicolumn{4}{c}{0.05}\\
        lr & 0.002 & 0.0005 & 0.0005 & 0.0005 \\
        layer\_decay & 0.75 & 0.75 & 0.75 & 0.75  \\
        warmup\_lr & 1e-6 & 1e-8 & 1e-6 & 1e-6  \\
        min\_lr & 1e-6 & 1e-5  & 1.5e-4 & 1.5e-3 \\
        warmup\_epochs & \multicolumn{4}{c}{5}\\
        
        \midrule \textit{augmentations} \\
        color\_jitter & \multicolumn{4}{c}{0.4} \\ 
        horizontal\_flip & True & True & False & False\\
        num\_sample & \multicolumn{4}{c}{2} \\
        aa & \multicolumn{4}{c}{rand-m7-n4-mstd0.5-inc1} \\
        smoothing & \multicolumn{4}{c}{0.1} \\
        train\_interpolation & \multicolumn{4}{c}{bicubic} \\
        test\_num\_segment & 5 & 5 & 2 & 2 \\
        test\_num\_crop & 3 & 3 & 3 & 3 \\
        
        \midrule \textit{erase} \\
        prob & \multicolumn{4}{c}{0.25}\\
        mode & \multicolumn{4}{c}{pixel}\\
        count & \multicolumn{4}{c}{1}\\
        split & \multicolumn{4}{c}{False}\\
        
        \midrule \textit{mixup} \\
        mixup & \multicolumn{4}{c}{0.8} \\
        cutmix & \multicolumn{4}{c}{1.0}\\
        mixup\_prob & \multicolumn{4}{c}{1.0}\\
        mixup\_switch\_prob & \multicolumn{4}{c}{0.5}\\
        mixup\_mode & \multicolumn{4}{c}{batch} \\
        \bottomrule
    \end{tabular}}
\end{table}

\section{Extra Results} \label{app:extra_results}

\subsection{Frozen Evaluation.}

\begin{table}[h]
    \centering
    \caption{\small{\bf Linear vs. Attentive Probe Evaluation for \putalg and VideoMAE.} We evaluate the effect of linear (Lin.) and attentive (Att.) probing when adapting \putalg to the K400 ($16\times 5\times 3$) and SSv2 $(16\times 2\times 2)$ tasks. \putalg and VideoMAE benefit from using a non-linear attentive probe.} 
    \label{tb:lin_att_video}
    {\fontsize{8.5pt}{8.5pt}\selectfont
    \begin{tabular}{l l c c c c}
        & & \multicolumn{2}{c}{\bf K400}  & \multicolumn{2}{c}{\bf SSv2}\\
        \bf Method & \bf Arch. & Lin. & Att. & Lin. & Att.\\
        \toprule
        VideoMAE  & ViT-L/16 & 52.5 & 77.8 & 41.3 & 61.2\\
        V-JEPA & ViT-L/16 & \cc 56.7 & \cc \bf 80.8 & \cc 50.1 & \cc \bf 69.5\\
    \end{tabular}}
\end{table}

\begin{table}[h]
    \centering
    \caption{\small{\bf Linear vs. Attentive Probe Evaluation for DINOv2 and OpenCLIP.} We evaluate the effect of linear (Lin.) and attentive probing (Att.) when adapting DINOv2 and OpenCLIP. Image-baselines benefit from using an attentive probing strategy. Results shown in \colorbox{gray!20}{gray} are reported from the linear probe evaluation in~\citet{oquab2023dinov2}.} 
    \label{tb:lin_att_image}
    {\fontsize{8.5pt}{8.5pt}\selectfont
    \begin{tabular}{l l c c c c c c c c c c }
        & & \multicolumn{2}{c}{\small\bf K400} & \multicolumn{2}{c}{\small\bf SSv2} &\multicolumn{2}{c}{\small\bf IN1K}  & \multicolumn{2}{c}{\small\bf Place205} & \multicolumn{2}{c}{\small\bf iNat21}\\
        \bf\small Method & \bf\small Arch. & \small Lin. & \small Att. &  \small Lin. & \small Att. & \small Lin. & \small Att. & \small Lin. &  \small Att. & \small Lin. &  \small Att.\\
        \toprule
        DINOv2 & ViT-g/14 & \ccg 78.4 & 83.4 & \ccg 38.3 & 50.0 & \ccg 86.5 & 86.2 & \ccg 67.5 & 68.4 & \ccg 85.7 & 88.8\\
        OpenCLIP & ViT-G/14 & \ccg 78.3 & 81.8 & \ccg 35.8 & 34.8  &\ccg 86.2 & 85.3 & \ccg 69.8  & 70.2 & \ccg 76.0 & 83.6\\
    \end{tabular}}
\end{table}

\paragraph{Linear vs. Attentive probe}
Table~\ref{tb:lin_att_video} shows that \putalg and VideoMAE benefit from using a non-linear attentive probe and multiple clips on the K400 and SSv2 downstream tasks.
Additionally, Table~\ref{tb:lin_att_image} shows that attentive probing leads to better performance on average for DINOv2 and OpenCLIP models.
Since attentive probing and multiclips eval improves the performance of all models, we use it as our default protocol in frozen evaluation.

\begin{table}[h]
    \centering
    \caption{\small{\bf Temporal Coverage on Kinetics-400.} We evaluate the effect of temporal coverage on K400. We train an attentive probe on K400 using either 1 clip ($\approx$ 2 seconds of a video) or 8 clips ($\approx$ 16 seconds of a video). To sample $N$ clips, we first divide a video in $N$ equal-length temporal segments and sample one clip at random per segment. The video encoder processes each clip in parallel and all the encoder output tokens are concatenated at the input of the attentive probe. Increasing the temporal coverage from 1 clip per video to 8 clips significantly improves the performance for both our VideoMAE baseline and \putalg.}
    \label{tb:lin_nbclips}
    {\fontsize{8.5pt}{8.5pt}\selectfont
    \begin{tabular}{l l c c}
        \bf\small Method & \bf\small Arch. & 1 Clip & 8 Clips\\
        \toprule
        VideoMAE & ViT-L/16 &  69.4 & 77.8\\[1ex]
        \multirow{1}{*}{V-JEPA} & ViT-L/16 & \cc 73.7 & \cc 80.9\\
    \end{tabular}}
\end{table}

\paragraph{One Clip vs Multiple clips.}
We examine the impact of changing the temporal coverage of a model during downstream evaluation on K400 action classification.
In Table~\ref{tb:lin_nbclips}, we evaluate VideoMAE and \putalg models using an attentive probe with access to either the feature map of 1 clip randomly sampled from the video, or the concatenated feature map of 8 clips randomly sampled from the video.
To sample 8 clips from a video, we first divide the video into 8 equal length temporal segments, and sample 1 clip at random from each segment.
A single clip corresponds to $\approx$ 2 seconds of a video on average, while 8 clips correspond to $\approx$ 16 seconds.
The video encoders processes each clip separately to produce a clip-level feature map, which are then concatenated at the input to the attentive probe.

Increasing the temporal coverage from 1 clip per video to 8 clips improves the performance of both \putalg and VideoMAE on K400 action classification.
We therefore use the multiclip attentive probing setup as our default evaluation pipeline.

\begin{table}[!h]
    \centering
    \caption{\small{\bf Finetuning results.} We evaluate a V-JEPA model with the finetuning protocol on the K400 and SSv2 datasets using 16 frames per clip and multi-view fusion (5$\times$3 or $2$×$3$) for inference. The {\bf \#Samples Seen} entry corresponds to the number of video clips processed during pretraining, which is larger than the size of the pretraining dataset for multi-epoch training. We compare \putalg with different video self-supervised learning approaches.
    We report the VideoMAEv2 results without instruction-turning for consistency with the other approaches. \putalg obtains competitive performance using the finetuning protocol.}
    \label{tab:finetune}
    {\fontsize{8pt}{8pt}\selectfont
    \begin{tabular}{l l l l c c}
        \multirow{2}{*}{\bf Method} & \multirow{2}{*}{\bf Arch.} & \multirow{2}{*}{\bf Pretraining Data} & \multirow{2}{*}{\bf \#Samples Seen } & \bf K400 & \bf SSv2 \\ 
        &  &  &  & {\fontsize{5.5pt}{5.5pt}\selectfont(16$\times$5$\times$3)} & {\fontsize{5.5pt}{5.5pt}\selectfont(16$\times$2$\times$3)}\\
        \toprule
        \multirow{2}{*}{VideoMAEv1} & ViT-L/16 & K400$\lvert$SSv2 & 380M$\lvert$410M & 85.4 & 74.3\\
        &  ViT-H/16 & K400$\lvert$SSv2 & 380M$\lvert$410M & 86.6 & 74.8\\ 
        \multirow{1}{*}{VideoMAEv2} &  ViT-H/16 &  Un.Hybrid & 1600M & 86.9 & 76.8\\
        \multirow{2}{*}{MVD} &  ViT-L/16 &  K400+IN1K & 2400M &  86.4 & 76.7\\
        &  ViT-H/16 &  K400+IN1K & 2400M & \bf 87.2 & \bf 77.3\\
        \midrule
        \multirow{2}{*}{V-JEPA} & ViT-L/16 & VideoMix2M & 270M & 85.6 & 75.1\\
        & ViT-H/16 & VideoMix2M & 270M & 86.6 & 77.0\\
    \end{tabular}}
\end{table}

\subsection{Finetuning} \label{sec:app_finetune}
In Table~\ref{tab:finetune}, we evaluate \putalg using finetuning (separately) on K400 and SSv2. We compare \putalg  with VideoMAEv2~\citep{wang2023videomae}, VideoMAE~\citep{tong2022videomae} and MVD~\citep{wang2023masked} using a ViT-L/16 or a ViT-H/16 architecture. \putalg obtains competitive performance using a finetuning protocol.
With a ViTiH/16 architecture, \putalg outperforms by $1.2\%$ VideoMAE and $+0.3\%$ VideoMAEv2 on the SSv2 dataset, while obtaining comparable performance on K400. 
\putalg also obtains performance similar to MVD on the SSv2 dataset.
The MVD model achieves the best performance across models on the K400 dataset, and is trained using the image dataset ImageNet1K, in contrast to the other methods in the table, which only use video data.
Additionally MVD requires the processing of significantly more samples during pretraining due to the cost of training the teacher encoder networks in a pre-pre-training step.

\subsection{Sample Efficiency of pretraining}
\label{app:samples}

We compare the sample efficiency of pretraining various state-of-the-art image and video models.
Specifically, we look at the number of samples (image or video clips) processed by the network during pretraining, which is larger than the size of the pretraining dataset for multi-epoch training.
Notably, our results with \putalg are obtained while processing an order of magnitude fewer samples than previous methods, and notably two orders of magnitude fewer samples than OpenCLIP.
We believe that further investment towards improving the video pretraining data distribution could lead to substantial gains in downstream image and video tasks.

\begin{table}[h]
    \centering
    \caption{\small{\bf Sample efficiency.} We compare the sample efficiency of pretraining various state-of-the-art image and video models. The {\bf \#Samples Seen} entry corresponds to the number of samples (image or video clips) processed by the network during pretraining, which is larger than the size of the pretraining dataset for multi-epoch training. The \putalg results in this paper are obtained while processing an order of magnitude fewer samples than previous methods.}
    \label{tab:samples}
    {\fontsize{8.5pt}{8.5pt}\selectfont
    \begin{tabular}{l l l | r}
         \bf Method & \bf Arch. & \bf Data & \bf \#Samples Seen  \\\toprule
         OpenCLIP & ViT-G/14 & LAION-2B & 39000M \\
         DINOv2 & ViT-g/14 & LVD 142M & 1900M \\
         VideoMAEv2 & ViT-g/14 & UnlabeledHybrid & 1600M \\
         V-JEPA & ViT-H/16$_{384}$ & VideoMix2M & 210M
    \end{tabular}}
\end{table}

\subsection{Masking Strategy}
\label{app:masking_ablation}

\begin{figure}[t]
    \begin{subfigure}[b]{0.285\textwidth}
        \centering
        \includegraphics[height=3cm]{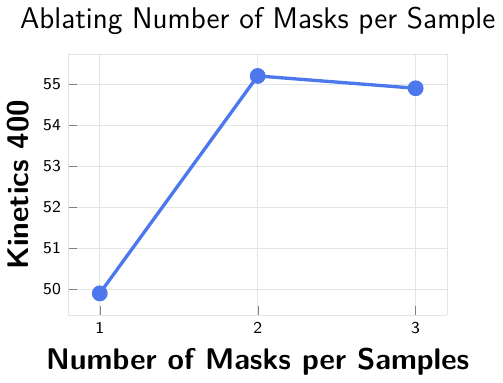}
        \caption{}\label{fig:masking_ablatio_num}
    \end{subfigure}
    \begin{subfigure}[b]{0.285\textwidth}
        \centering
        \includegraphics[height=3cm]{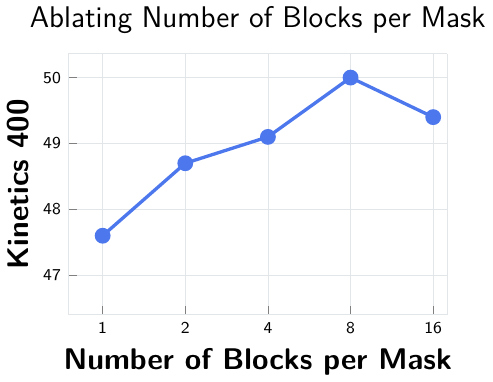}
        \caption{}\label{fig:masking_ablation_blocks}
    \end{subfigure}\hfill
    \begin{subfigure}[b]{0.42\textwidth}
        \centering
        \includegraphics[height=3cm]{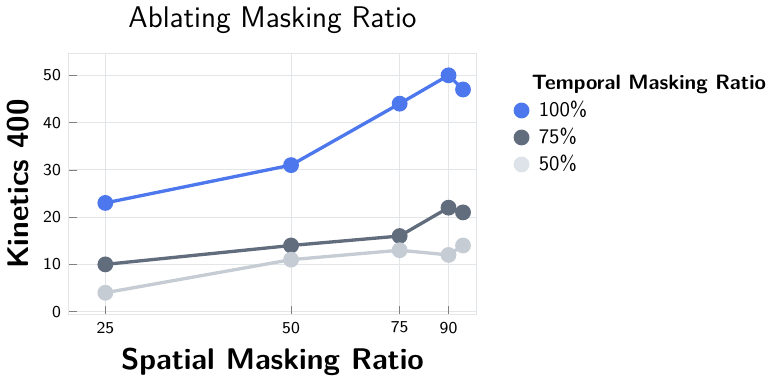}
        \caption{}\label{fig:masking_ablatio_ratio}
    \end{subfigure}
    \caption{\small {\bf Masking Strategy Ablation.} 
    Evaluating a linear probe on a ViT-B/16 pretrained with \putalg on K400 under various 3D Multi-Block masking settings.
    We examine the impact of {\bf (a)} sampling several masks per video, {\bf (b)} varying the number of blocks in a mask, and {\bf (c)} varying the average spatial and temporal masking ratio. A temporal masking ratio of 100\% extends the spatial mask across all the frames in the clip.
    We find it important to maintain a high spatial and temporal masking ratio during pretraining.}
    \label{fig:masking_ablation}
\end{figure}

An important component of the \putalg pretraining strategy is the 3D clip masking strategy. In this section, we detail 26 ablation experiments exploring different masks. For all the experiments, we pretrain a ViT-B/16 pretrained on K400. Figure~\ref{fig:masking_ablation} presents a summary of those results.

Figure~\ref{fig:masking_ablatio_ratio} shows the effect of changing the spatial and temporal masking ratio.
Figure~\ref{fig:masking_ablation_blocks} ablates the number of sampled blocks used to construct the masks given a fixed effective masking ratio of $90\%$.
Finally, in Figure~\ref{fig:masking_ablatio_num} we examine our multi-masking strategy and find that sampling two masks for each clip (long-range and short-range) to be more effective than sampling just a single mask for each clip.

In Figure~\ref{fig:masking_ablatio_ratio}, we explore different average spatial and temporal masking ratio, i.e. the spatial/temporal ratio of the area that is covered by a mask on average for a clip. Recall that each mask is constructed by sampling several (possibly overlapping) blocks and taking their union.
We change the average spatial or temporal masking ratio by changing a block spatial or temporal size, as well as the overall number of blocks. We found that low spatial or temporal coverage results in a trivial prediction task, which degrades downstream performance. Based on those results, we sample masks that remove roughly $90\%$ of the frame and extend along the entire temporal dimension of the clip by default.

\begin{figure}[t]
  \centering
  \begin{subfigure}[b]{0.49\textwidth}
    \centering
    \includegraphics[width=1\linewidth]{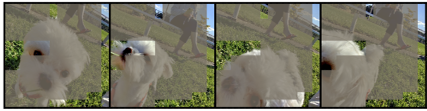}
    \caption{Num. Blocks: 8, Spatial Block Size: $32 \times 32$}
  \end{subfigure}\\
  \begin{subfigure}[b]{0.49\textwidth}
    \centering
    \includegraphics[width=1\linewidth]{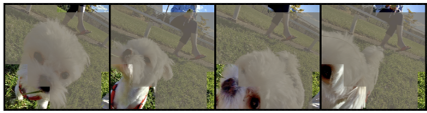}
    \caption{Num. Blocks: 4, Spatial Block Size: $80 \times 80$}
  \end{subfigure}
  \begin{subfigure}[b]{1\textwidth}
    \centering
    \includegraphics[width=0.49\linewidth]{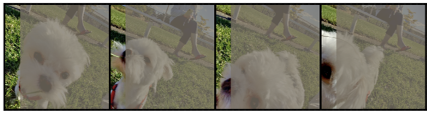}
    \caption{Num. Blocks: 2, Spatial Block Size: $160 \times 160$}
  \end{subfigure}
    \caption{Illustration of mask with number of blocks and  block size. Each mask is constructed by sampling several (possibly overlapping) blocks and taking their union.}
    \label{fig:masks_ablation_visualizaiton}
\end{figure}

In Figure~\ref{fig:masking_ablation_blocks} , we explore different block size given an effective spatial masking ratio of  90\%  and temporal ratio of 100\%. We keep the masking ratio approximately constant by changing the block size and the number of block at the same time. We find that sampling several blocks to perform better than sampling a single large block. Figure~\ref{fig:masks_ablation_visualizaiton} visually illustrates the effect of sampling several smaller blocks to construct a mask.

In Figure~\ref{fig:masking_ablatio_num}, we explore the effect of sampling various number of masks per samples. We find that sampling two masks for each clip, with different spatial block sizes for each, to be more effective than sampling just a single mask.  We hypothesize that this masking strategy induces complementary tasks. In our experiment, we use this as our default masks sampling.

\end{document}